\documentclass[MS]{macro/neu_msthesis}

\usepackage{physics}
\usepackage{siunitx}
\usepackage{graphics} % for pdf, bitmapped graphics files
\usepackage{graphicx} % for epx
\usepackage{epsfig} % for postscript graphics files
\usepackage{lipsum} 
\usepackage{amsmath, bm} % assumes amsmath package installed
\usepackage{amssymb}  % assumes amsmath package installed
\usepackage{upgreek}
\usepackage{url}
\usepackage{breakurl}
\usepackage{caption}
\usepackage{subcaption}
\usepackage{pgfplots}
\usepackage{floatrow}
\usepackage{float}
\usepackage[ruled,linesnumbered]{algorithm2e}
\usepackage{multicol}
\usepackage{tikz}
\usetikzlibrary{shapes, arrows}
\usepackage{amsmath}
\newfloatcommand{capbtabbox}{table}[][\FBwidth]

\title{Towards Autonomous Multi-Modal Mobility Morphobot (M4) Robot: Traversability Estimation and 3D Path Planning}

\author{Rohit Hiraman Rajput}

\dept{Mechanical and Industrial Engineering}

\degree{Master of Science}

\degreename{Robotics}

\field{Robotics}

\submitdate{August 2023}

\numberofmembers{3}

\principaladviser{Dr. Alireza Ramezani, Dr. Hanumant Singh}
\firstreader{Dr.Rifat Sipahi}
\chairman{Dr. Marilyn Minus}

\dean{Sagar Kamarthi}

\usepackage{amsfonts}			% special math characters
\usepackage{times}		% use Postscript fonts

\newcommand{\secref}[1]{Section~\ref{#1}}

\newcommand{\ifno}[1]{}

\newcommand{\etal}{\emph{et~al.}}

\usepackage{multirow}
\makeindex
\usepackage{makeidx}
\usepackage{acronym}

\usepackage{url}
\usepackage{graphicx}
\ifnum\pdfoutput>0
\usepackage[pdftex,%                    % hyper-references for ps2pdf
bookmarks=true,%                        % generate bookmarks ...
bookmarksnumbered=true,%                % ... with numbers
hypertexnames=false,%                   % needed for correct links to figures
breaklinks=true           %Allows link text to break across lines;
]{hyperref}
\else
\usepackage[hypertex,%                  % hyper-references for ps2pdf
bookmarks=true,%                        % generate bookmarks ...
bookmarksnumbered=true,%                % ... with numbers
hypertexnames=false,%                   % needed for correct links to figures
breaklinks=true           %Allows link text to break across lines;
]{hyperref}
\fi

\hypersetup{				% setup PDF information fields
pdfauthor = {\authorRef},
pdftitle = {\titleRef},
pdfsubject = {\expandafter{\degreeRef} thesis submitted to Northeastern University},
pdfkeywords = {Multimodal robots,M4,Traversability,3D path planning}
pdfcreator = {LaTeX with hyperref package},
pdfproducer = {dvips + ps2pdf}}

\begin{document}

% add a pdf bookmark to the cover page
\pdfbookmark[1]{Cover}{cover}

% --- title page ---
\titlepage

% --- front matter ---
\begin{frontmatter}
% print signature page
%\signaturepage
% dedication
%\input{tex/dedication.tex}

% table of content (add bookmark for convenience)
\pdfbookmark[1]{Table of Contents}{contents}
\tableofcontents
\listoffigures
\newpage\ssp
\listoftables

% include a list of Acronyms (comment out if no acronyms are specified)
%\input{tex/acronyms.tex}

% include any of the front matter files that contain text
% attention the input does cause a page break, the include on 
% the other hand does not
% acknowledgements.tex:

\begin{acknowledgements}

I extend my heartfelt gratitude to all the members of the SiliconSynapse Lab and Nufr Lab at Northeastern University for their unwavering support throughout my thesis journey. I am especially grateful to Prof. Alireza Ramezani and Prof. Hanumant Singh, my primary and co-advisers, respectively, for their exceptional mentorship, belief in my abilities, and invaluable guidance. Their expertise and dedication have inspired me to strive for excellence in my research. I would also like to acknowledge the contributions of Dr. Milad Ramezani (CSIRO) and Adarsh Salagame, whose support and collaboration have enriched my work. Special thanks go to Prof. Rifat Siphai for serving as my mechanical engineering department co-adviser and thesis reader. I am deeply thankful for the support of my parents, whose unwavering love and encouragement have been the driving force behind my achievements. I am excited about the future possibilities it opens up in the field of robotics and autonomous systems.

\end{acknowledgements}

% abstract.tex:

\begin{abstract}

This thesis enhances the autonomy of the M4 (Multi-Modal Mobility Morphobot) robot, designed for Mars and rescue missions. The research enables the robot to autonomously select its locomotion mode and path in complex terrains. Focusing on walking and flying modes, a Gazebo simulation and custom perception-navigations pipelines are developed. Leveraging deep learning, the robot determines optimal mode transitions based on a 2.5D map. Additionally, an energy-efficient path planner based on 2.5D mapping is implemented and validated in simulations. The contributions demonstrate scalability for future mode integrations. The M4 robot showcases intelligent mode switching, efficient navigation, and reduced energy consumption, bringing us closer to fully autonomous multi-modal robots for exploration and rescue missions. This work paves the way for future advancements in autonomous robotics, with the ultimate vision of deploying the M4 robot for exploration and rescue tasks, making a significant impact in the quest for intelligent and versatile robotic systems.

\end{abstract}

\end{frontmatter}

% --- body of the document ---

%\pagestyle{plain}
\pagestyle{headings}

% include each chapter like below
% intro.tex:
\chapter{Introduction}
\label{chap:intro}

\section{Background and Motivation}

This Thesis aims to add autonomy to the M4 (Mars Multi-Modal Morphofunctional) robot's multi-Modal capabilities to choose its mode and its path in uneven or complex terrain. The robot is recently developed at Caltech Aerospace CAST (Center for Autonomous Systems and Technologies). The guiding vision is to ultimately develop a system for the M4 to know when to switch between multiple forms to traverse the uneven or complex terrain and create an optimal path for the robot considering the energy consumption for each mode
all by itself without relying on GPS, making use of its morphing abilities to switch between locomotion modes in an optimal fashion.
In the early stages of the project, it became clear that such ambitious tasks would require more work than could possibly be accomplished in the eight months allocated to the thesis. In fact, the challenge include major actively researched problems such as the eight-multimodal capacities determination for determining when to morph into a locomotion mode or Segway mode in an energy-efficient way , therefor the focus has been shifted to the subset of the above. the goal of this work is to put together a fully working model in simulation of the robot and create a whole navigation and perception pipeline for only 2 modes of the M4 locomotion and flight mode respectively with scalability for other modes in the future and solve the research problem of determination of when to morph between the modes using deep learning and also create an optimal path planner based on 2.5D map for getting a real-time operation.

This thesis presents a comprehensive investigation into enhancing the autonomy of the M4 (Mars Multi-Modal Morphofunctional) robot, a state-of-the-art creation developed at Caltech Aerospace CAST (Center for Autonomous Systems and Technologies). The primary objective of this research is to endow the M4 robot with the ability to independently select its locomotion mode and path in uneven or complex terrains while considering energy efficiency. The overarching vision is to achieve a system where the M4 robot intelligently transitions between multiple locomotion modes and self-optimizes its path, leveraging its morphing capabilities in a sophisticated manner.
 
\begin{figure}[!h]
\centering\includegraphics[width=0.7\textwidth]{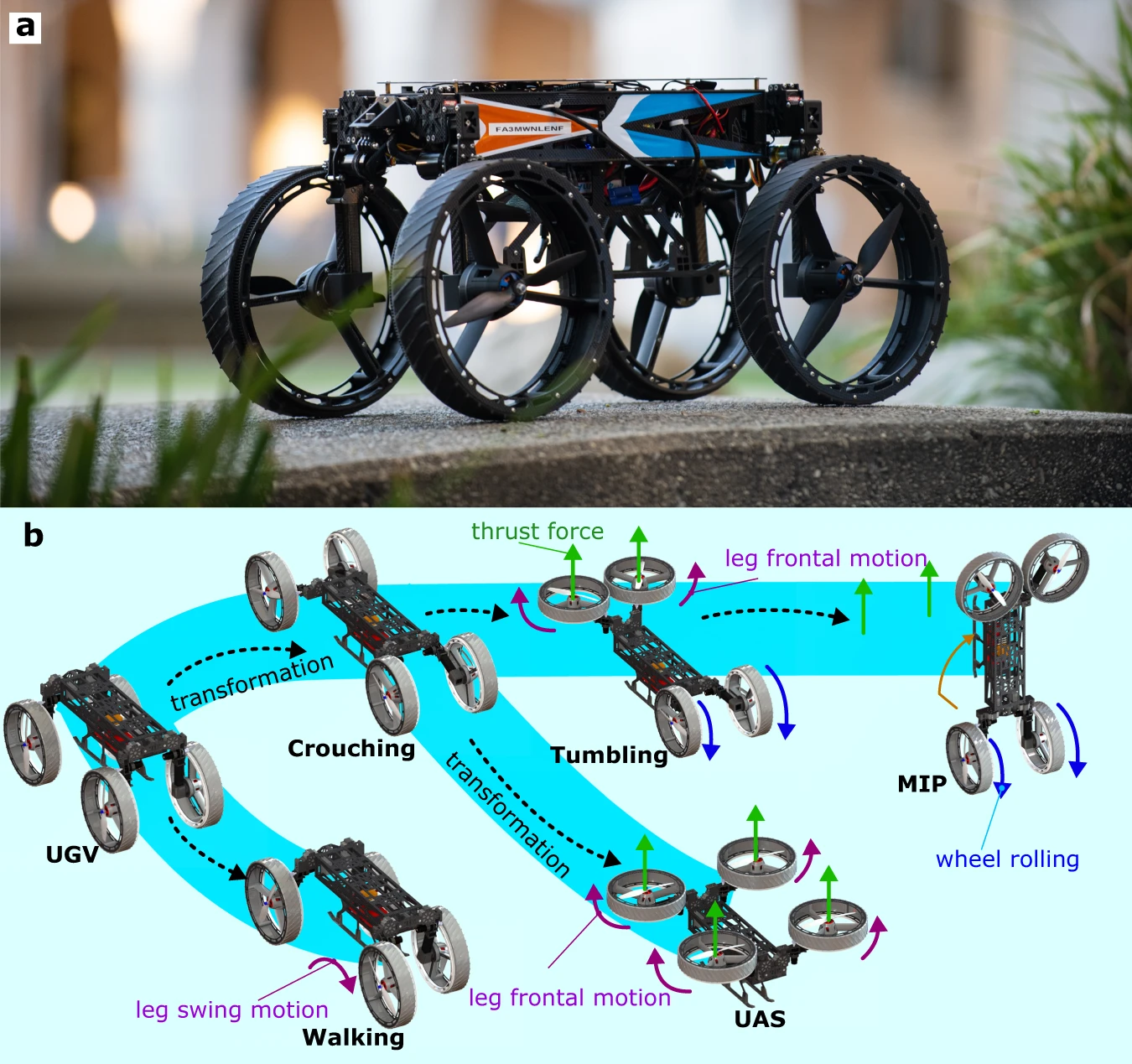}
\caption{a Shows M4 in wheeled mode. b Illustrates cartoon depictions of M4’s transformation to other modes.\cite{M4_nature}.}
\label{fig:m44}
\end{figure}

The early stages of the project underscored the enormity of the challenge at hand, particularly in dealing with the intricacies of determining six multimodal morphing capacities \ref{fig:m44} and achieving energy-efficient morphing. Consequently, the project's scope was thoughtfully adapted to concentrate on specific aspects while paving the way for future expansion. The immediate goal was to develop a fully functional simulation model of the M4 robot and create a navigation and perception pipeline, tailored initially to two modes of locomotion - walking and flying. The proposed approach ensures scalability for accommodating additional modes in the future, thereby facilitating the eventual realization of the M4 robot's complete multi-modal capabilities.

This thesis presents a comprehensive solution that capitalizes on existing deep learning models to address the research problem of determining when to morph between the two chosen locomotion modes. By leveraging the power of pre-existing deep learning algorithms, the M4 robot can efficiently adapt to varying terrain conditions in real-time. Extensive datasets encompassing diverse terrains are utilized to train the deep learning model for wheel mobility, facilitating the robot's ability to make informed decisions regarding mode selection while navigating through complex landscapes.

Furthermore, a cutting-edge optimal path planner, based on a 2.5D map representation of the environment, has been devised to enable the M4 robot to chart its course with utmost efficiency while considering energy consumption and less processing. The path planner's effectiveness was validated through 3 different scenarios in simulations, showcasing its ability to perform in real-time operations given the static 2.5D map and generate the optimal path.

The culmination of this research has seen the successful implementation and validation of the autonomous multi-modal control for the M4 robot in a simulated environment. The robot demonstrated its capability to traverse complex terrains by intelligently switching between wheeled and flying modes while efficiently conserving energy. These achievements represent significant milestones towards the overarching vision of creating a fully autonomous system for the M4 robot, empowering it to explore and navigate diverse environments without relying on GPS with all of its morphing modes.

In conclusion, this thesis serves as a solid foundation for future advancements in the realm of multi-modal robotics and autonomous systems. The proposed deep learning-based mode determination and optimal path planning techniques exhibit promising potential for seamless integration with additional locomotion modes and more complex terrains. As the M4 robot moves towards real-world deployment, this research opens up exciting possibilities for a new generation of autonomous robots that can adapt, learn, and explore the unknown frontiers with unparalleled intelligence and efficiency.

The resulting proof of concept in simulation and with real-world data from realsense camera is designed to be a significant step toward adding full morphing configurations of the robot on the real robot. In line with this reasoning, this report is meant as an integral part of the project documentation, providing explanations and suggestions that have not necessarily made it to the code repository or to the shared drive.

\begin{figure}[!h]
\centering\includegraphics[width=0.7\textwidth]{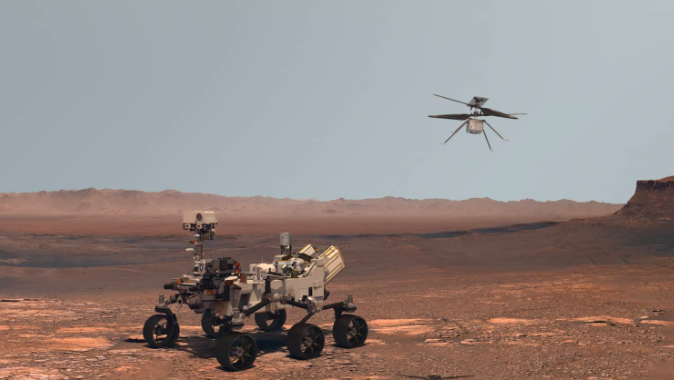}
\caption{Artist rendering of the Perseverance rover and Ingenuity helicopter exploring Mars \cite{marsbots}.}
\label{fig:marsbots}
\end{figure}

\section{Past Work and History of the M4 Robot}

\label{M4}

In recent years, the fascination with space exploration, particularly concerning Mars, has sparked a surge in innovative technologies and new challenges. NASA and JPL, building upon the successes of the Spirit, Opportunity, and Curiosity rovers, have been eagerly exploring ways to enhance Martian exploration. With the introduction of more complex and heavier platforms, such as the Perseverance rover carrying the Ingenuity helicopter and an array of scientific instruments, see Figure \ref{fig:marsbots}, the cost of Mars missions has been on the rise. However, the potential of these robots remains limited by the harsh and challenging terrains they must traverse.\cite{M4_nature}

The journey towards developing multi-modal robots began with thruster-assisted legged mobility for efficient Mars exploration \cite{ramezani2022thruster}. This work was extended to optimize ground contact forces in quadrupedal locomotion \cite{sihite2021unilateral}, ensuring stable gaits. Integration of thrusters into bipedal legged locomotion led to Harpy, enabling controlled thruster-assisted mobility \cite{dangol2021control,dangol2021reduced,dangol2020performance,dangol2021hzd,dangol2020NEW,dangol2020thruster,de2020thruster,dangol2020feedback}. This evolved into the Husky Carbon platform, a multi-modal legged-aerial robot \cite{salagame2022letter,dangol2020towards}.

The concept of Morpho-functional robots, capable of both legged and aerial modes, was introduced \cite{ramezani2023morpho}. Innovative research enabled legged walking on inclined surfaces \cite{changaoss1}, while a scalable UGV-UAV morphing robot demonstrated autonomous 3D path planning \cite{sihite2023demonstrating}. Culminating in the Multi-Modal Mobility Unmanned Vehicle, principles of thruster-assisted legged and aerial mobility were combined \cite{ramezani2023multi}. This concept was substantiated by efficient aerial mobility designs for Mars exploration \cite{ramezani2022efficient}. Overall, these works showcase the progressive development of dynamic and adaptable multi-modal robots.

\begin{figure}[!h]
    \centering
    \includegraphics[width=0.9\textwidth]{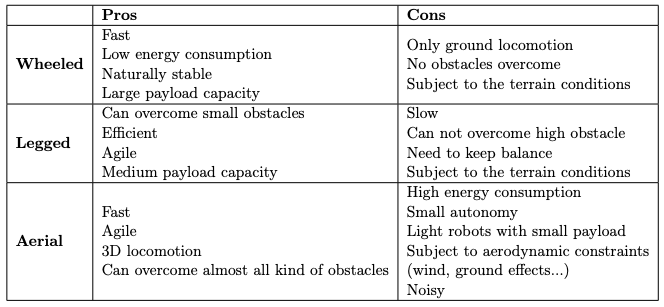}
    \caption{Various advantages and disadvantages of different locomotion modes \cite{ben}.}
    \label{fig:multimode}
\end{figure}

\begin{figure}[!h]
    \centering
    \includegraphics[width=0.8\textwidth]{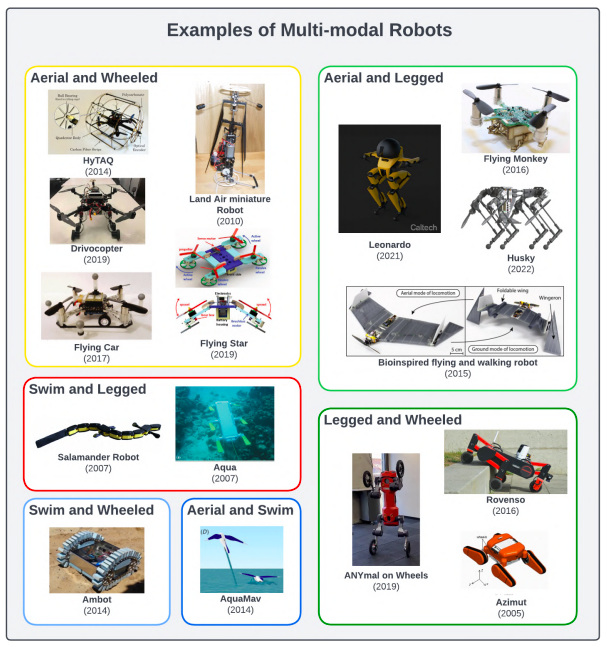}
    \caption{Examples of multi-modal robots \cite{MM1, MM2, MM3, MM4,ANYmal_Wheeled,salamandre,PRM-MM, flying_monkey, Flying_star,kim2021bipedal}.}
    \label{fig:multimodal}
\end{figure}

Building upon this groundwork, the idea of the M4 robot emerged, driven by a vision to harness the advantages of multiple locomotion modes for exploration. One of the remarkable aspects of living creatures is their ability to adapt to varying scenarios and environments by utilizing their bodies in versatile ways. Conventional robots struggle to replicate such adaptability. The M4's ingenious bio-inspired design seeks to integrate the qualities of wheeled, legged, and aerial robotics into a single platform, see Figure \ref{fig:multimode}. Like the other robots showcased in Figure \ref{fig:multimodal}, the M4 boasts diverse locomotion modes, but what sets it apart is its capability to morph its body, unlocking an even broader range of properties. 

Despite the increased complexity of a morphing robot, it's compact size and integration of multiple locomotion modes make the M4 a compelling candidate for future missions. For instance, while a traditional rover might need to deploy a helicopter to investigate its surroundings, the M4 could independently fly its instruments in and out of a crater. Furthermore, this cutting-edge technology could prove invaluable in search and rescue operations, where a morphing and multi-modal body may be essential for accessing and navigating tight spaces.\cite{M4_nature}

\begin{figure}[!ht]
\centering\includegraphics[width=\textwidth]{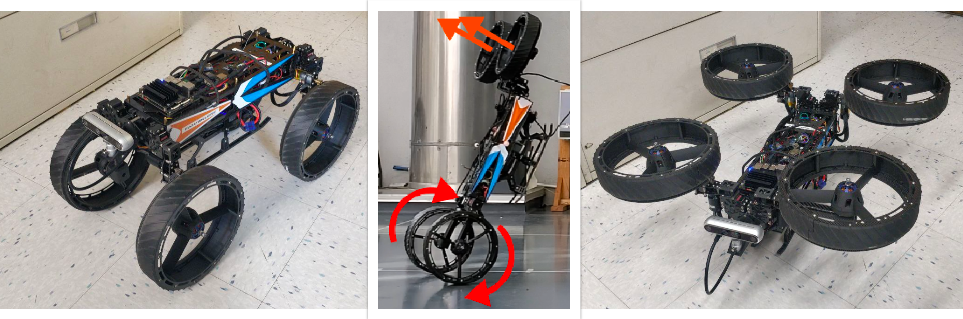}
\caption{The M4 robot in various morphing configurations: rover, segway, drone.\cite{filip}}
\label{fig:morph}
\end{figure}

The morphing capabilities of the M4 make it possible to traverse a wider range of terrains more efficiently than its unimodal counterparts.\cite{M4_nature} It can drive like a rover, stand like a segway and fly like a drone, see Figure \ref{fig:morph}, and more is yet to be imagined. The current robot is still the first version, a prototype put together in record time. Therefore, some mechatronic design issues are still unknown and expected but that shouldn't stop us from working on its autonomy.\cite{ben} The main point here is that the robot was not exactly designed to accommodate additional hardware, its body is already completely packed so new components have to be mounted on the outside.\cite{filip}

Previously, Eric Sihite and Benjamin Mottis\cite{ben} played pivotal roles in bringing the M4 robot to life and developing a path-planning algorithm that optimizes its morphing and multimodal capabilities in an energy-efficient manner \cite{PRM-MM}. also, Filip Slezák\cite{filip} integrated the rtabmap localization on the robot and worked on 2d navigation planning on board with jetson nano which paved the way for this research to operate on the real robot.\cite{filip}\cite{M4_nature}

\section{Contributions and Thesis Outline}

This thesis makes notable contributions towards enhancing the autonomy of the M4 robot in multi-modal locomotion and flight control for complex environments. The key contributions can be summarized as follows:

\begin{itemize}
   \item Efficient 3D Path Planning with Energy Considerations through 2.5D mapping: The thesis proposes an optimal path planner based on a 2.5D map representation of the environment. This planner empowers the M4 robot to chart its course in a manner that minimizes energy consumption while ensuring efficient navigation through challenging terrains for multimodal robots. The validation of the path planner through different scenarios for multimodal robot simulations demonstrates its effectiveness for using multiple modes of the robot showcasing its potential as a critical component of the M4 robot's autonomy.
   
    \item Custom Perception and Navigation Pipeline: A critical aspect of the research involves designing and implementing a custom perception and navigation pipeline tailored to the unique requirements of the M4 robot. This pipeline encompasses various combinations of elevation mapping and traversability with the navigation system directly for Multimodal systems. The novel combination of perception and navigation elements ensures that the M4 robot can effectively sense and understand its environment, thus making informed decisions during mode transitions and path planning
        
    \item Autonomous Mode Determination Using Existing Deep Learning Models: The research successfully addresses the challenge of determining when the M4 robot should morph between its walking and flying modes to traverse uneven or complex terrains efficiently. By leveraging existing deep learning models, the robot can make decisions on mode selection, enabling it to adapt intelligently to varying environmental conditions based on a 2.5D map.
   
    \item Multi-Modal Simulation in Gazebo: One of the key contributions is the successful creation of a comprehensive simulation environment for the M4 robot using Gazebo, a widely-used robotics simulation platform. The development of this simulation framework enables the testing and evaluation of the M4 robot's autonomy algorithms in diverse and complex terrains without the constraints of real-world experimentation. The Gazebo simulation provides a realistic and immersive environment, which proves crucial in validating the proposed autonomous control mechanisms for the M4 robot.

    \item Scalability for Future Multi-Modal Enhancements: The thesis design and implementation demonstrate scalability, allowing for the seamless integration of additional locomotion modes into the M4 robot's autonomy framework in the future. This adaptability ensures that the research findings can be extended to include a wide range of locomotion modes.
\end{itemize}

% --- EOF ---

% Path planning
\chapter{Literature Review}
\label{chap:Literature Review}

\section{Multimodal locomotion Path Planners}
\label{sec:multimodal}
In terms of global path planning optimization, numerous methods and solutions have been developed for rolling, walking, or flying robots. However, the domain remains relatively underexplored concerning multi-modal robots, presenting ample opportunities for further research. A common approach involves combining a discretization algorithm with a search algorithm to create a graph representation of the free space, followed by a post-process step to optimize the final path.

To map the environment and path planning,  Sharif \etal \cite{sharif_energy_2018} utilize a 3d uniform grid-based method and a Araki \etal \cite{araki_multi-robot_2017} also utilize a 3d uniform grid-based method but it has fixed a flight height for the robot. On the other hand, sampling methods are employed in \cite{suh_optimal_2019, composite_graph_optimization} to generate a graph-based representation of space for robots that can fly and do wheeled motion. The current state-the-art for multimodal robots path planner is Eric Sihite et al.\cite{PRM-MM} which uses a 3d MM-prm method to discretize the environment and sample it. To compute costs on the graph's edges, most articles consider factors such as power consumption \cite{sharif_energy_2018, araki_multi-robot_2017}. Suh \etal \cite{suh_optimal_2019} propose an optimization method using machine learning based on a reduced physical dynamics model. The optimization function accounts for constant power drain, battery voltage, and motor torque constant. Additionally, in the work by \cite{anymal_lune}, path costs are computed by formulating risk in various scenarios.also in the work for 3d For graph searching, Dijkstra's Algorithm \cite{dijkstra1959note} or its improved version, A$^\star$, is commonly implemented in most studies. A post-processing step involving a smoothing algorithm \cite{suh_optimal_2019} is often applied to optimize the path.

Other methods based on optimization have been explored for multi-modal path planning problems. For example, in \cite{composite_graph_optimization}, the authors employ the Trajectory-Equilibrium-based (TEB) optimization technique to improve an initial path generated by PRM and a search algorithm. This algorithm is used to optimize the energy consumption of a Vertical Take-Off and Landing (VTOL) drone operating in a constrained environment, where the robot switches between fixed-wing and quadcopter modes based on available free space.

\section{Mapping,Traversability and Navigation framework}

The literature review showcases the diversity of mapping techniques for both 3D and 2.5D environments in the context of autonomous robotic navigation. 3D mapping approaches, such as Occupancy Grid Mapping \cite{occupancy} and OctoMap \cite{octomap}, provide detailed representations of complex environments but come with high computational costs and memory requirements and are used in state-of-the-art multimodal planners. In contrast, 2.5D mapping techniques, represented as elevation maps, offer a compromise between computational complexity and information representation, making them suitable for resource-constrained robotic platforms. Works like robot-centric elevation mapping \cite{Fankhauser2018ProbabilisticTerrainMapping} have demonstrated the effectiveness of 2.5D mapping for robots in outdoor environment.
the traversability analysis and 2.5 mapping play a vital role in navigation, the current state of the art is Takahiro Miki \etal \cite{mikielevation2022} which uses cupy for fast embedded processing and CNN for traversability estimation, another paper P. Ewen \etal \cite{ewen2022these} creates a Sementic elevation map from camera as a semantic Bayesian inferencing framework for real-time elevation mapping and terrain property estimation.
Simultaneous Localization and Mapping (SLAM) techniques with cameras,state-of-the-art such as RTAB-Map \cite{RTABMAP} and ORB-SLAM3 \cite{campos2021orb}, play a vital role in real-time mapping and localization for both 3d and 2.5D mapping. Zhuozhu Jian \etal\cite{jian2022putn} shows a navigation framework for ground robots where it uses 2.5 mapping.

\section{Research analysis}

In this section, we identify critical research gaps in existing multimodal planners and mapping techniques for robotic systems with both flying and ground-based locomotion capabilities. The gaps highlight the limitations of current approaches and the novel directions our research aims to pursue.

\begin{itemize}
% addd section 1 review  callback command add 
\item A key gap in the existing multimodal planners mentioned in \secref{sec:multimodal} is the assumption of constant ground height for mode assignments, which fails to account for uneven terrains or even slightly elevated areas and steps. In reality, the constraints for ground height are dynamic, leading to inaccuracies and non-functionality of the planner in such environments. Addressing this gap is essential to ensure the planner's robustness and effectiveness in real-world scenarios.

% add some more here
\item Current multimodal planners predominantly rely on computationally expensive 3D mapping techniques for navigation, which might be impractical for onboard navigation due to limited computational resources. Surprisingly, there is a lack of research focusing on 2.5D mapping techniques specifically tailored for multimodal systems. This gap calls for the development of efficient and lightweight mapping strategies that can seamlessly accommodate the unique requirements of multimodal robots.

\item Moreover, existing mapping and navigation approaches tend to avoid obstacles from all sides, a standard practice suitable for aerial or mobile robots. However, this strategy might not be optimal for multimodal robots, which have the capability to land on top of obstacles and utilize less energy-consuming modes, resulting in improved energy efficiency. Bridging this gap entails designing adaptive obstacle avoidance algorithms that leverage multimodal capabilities to optimize energy consumption during navigation.

\end{itemize}

Our research addresses the identified challenges above by building upon Takahiro Miki \etal 's \cite{mikielevation2022} work to create a navigational pipeline for multimodal robots. We leverage their 2.5D mapping techniques to overcome the assumption of constant ground height, ensuring adaptability to uneven terrains. Additionally, we adopt a pre-trained CNN neural network for traversability estimation for ground locomotion robots which is based on Chavez-Garcia \etal's framework \cite{main_traversability}, enhancing computational efficiency while making informed decisions for energy optimization and autonomous mode selection. This integration paves the way for an innovative and energy-efficient navigation system for multimodal robots, contributing to the field of autonomous robotics and its applications in challenging environments.

% Methodology
% intro.tex:
\chapter{Methodology}
\label{chap :: Methodology}

The methodology chapter outlines the systematic approach undertaken to achieve the objectives of this thesis, which encompass the creation of a simulation model for the M4 robot and simulated environment in the gazebo and the development of a custom perception and navigation pipeline featuring a multimodal locomotion 3D path planner based on 2.5D mapping principles.

This chapter is organized as follows: Section 3.1 elaborates on the construction of the robot simulation model and environment in the gazebo, including its incorporation, integration of physics and dynamics, sensor simulation, and subsequent applications for testing and evaluation. Section 3.2 delves into the development of the custom perception and navigation pipeline, covering the whole custom pipeline with 2.5D environment elevation mapping, autonomous locomotion mode selection, and the incorporation of the Multimodal locomotion 3D path planner. Finally, Section 3.3 provides insights into the creation and integration of the multimodal locomotion 3D path planner, which serves as a crucial element in enhancing the robot's navigation capabilities.

By delving into the specifics of each phase, this chapter aims to offer a comprehensive understanding of the methodologies, algorithms, and tools used to accomplish the research goals. The subsequent sections provide a detailed account of the procedures and considerations involved in creating a sophisticated robot simulation model and designing a custom perception and navigation pipeline empowered by the 3D path planner based on 2.5D elevation mapping.

\section{Simulation Setup and Gazebo Environment}
\label{sec::simulation}

In pursuit of our research objectives, it's important to note that conducting experiments directly on a physical robot could entail potential risks, including damage to the robot itself, safety concerns, and costly iterations. Consequently, we adopted a simulation model to circumvent these challenges. This approach allowed us to systematically explore and validate the performance of the custom perception and navigation pipeline, as well as the 3D path planner, in a controlled and safe environment.
The simulation model was converted from its physical robot CAD to URDF to replicate the key physical attributes, kinematics, and sensor placements of the robot. Although virtual, this model allowed us to conduct controlled experiments and assessments that would otherwise be challenging or impractical with a physical counterpart.

Realistic physics and dynamics were seamlessly integrated into the simulation, providing a dynamic and responsive platform for testing and experimentation. This approach allowed us to examine the behavior of the robot within a controlled yet lifelike environment.

For the simulation setting, we chose Gazebo\cite{Gazebo}, a state-of-the-art open-source 3D robotics simulator as it stands out as a prime choice for the development and testing of robotic systems. This preference is underpinned by Gazebo's integration with the Robot Operating System (ROS) \cite{ROS} and also for our future exploration of deep learning training Gazebo is already proven effective.

Gazebo ROS provides an interface between the Gazebo simulator and the ROS middleware\cite{ROS}, enabling seamless communication between simulated robotic components and the software stack controlling them. This integration facilitates the testing of algorithms, control strategies, and perception techniques in a controlled and repeatable environment.

\begin{figure}[H]
    \centering
    \includegraphics[width=0.2\textwidth]{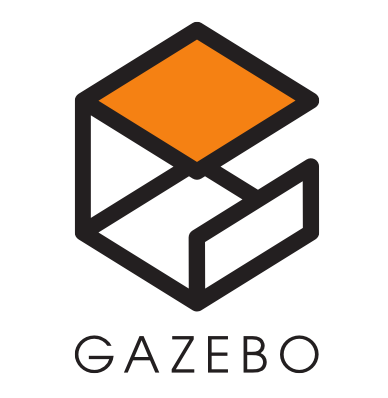}
    \caption{gazebo simulator logo \cite{Gazebo}} 
    \label{fig::gazebo_logo}
\end{figure}

some of the key features of the Gazebo :
\begin{itemize}
\item Seamless ROS Compatibility: Gazebo ROS seamlessly interfaces with the ROS middleware, creating a harmonious synergy that streamlines the simulation process. This compatibility ensures that simulated robotic components can be effortlessly integrated into the broader ROS ecosystem, easing the transition from virtual environments to real-world applications.

\item Immersive Realism: Gazebo ROS excels in providing a highly realistic simulation environment, equipped with advanced physics engines and rendering capabilities. The fidelity of these virtual environments is critical for testing and validating a wide spectrum of robotic scenarios, encompassing tasks such as navigation through complex terrains, object interaction, and sensor data interpretation.

\item Accurate Sensor Emulation: Gazebo ROS shines in its capability to accurately emulate diverse sensors, including cameras, lidars, and IMUs. This feature is pivotal for refining perception algorithms and sensor fusion techniques before deployment. By enabling the accurate reproduction of sensor data, Gazebo ROS facilitates the optimization of object detection, mapping, and localization strategies.

\item Modular Extensibility: Gazebo ROS's modular architecture, built around a plugin system, offers developers the freedom to tailor and expand the simulator's capabilities. This extensibility empowers the integration of customized controllers, sensor models, and environment components, enabling the fine-tuning of simulations to cater to diverse robotic applications.

\item Time and Cost Efficiency: By offering a controlled and repeatable testing environment, Gazebo ROS accelerates the development lifecycle of robotic applications. This efficiency translates to considerable time and cost savings, as developers can iteratively refine algorithms and software before committing resources to physical robot deployment.
\end{itemize}

\subsection{M4 simulated model}

During the development of the simulated M4 robot model, the initial step involves utilizing the original Solidworks CAD design, which encompasses the various components and camera placements of the robot. This design serves as the starting point, and its integration into the simulation is streamlined through the application of a Solidworks to URDF converter plugin. Coordinated systems are established for the robot's joints and links during this phase, which lays the foundation for ensuring accurate simulation outcomes. Subsequently, the CAD model is transformed into a URDF file, which serves as a critical input for the ensuing stages of the simulation process. This streamlined procedure underscores the meticulous conversion efforts undertaken to facilitate a seamless transition from the design to the simulation stage.

\begin{figure}[H]
    \centering
    \includegraphics[width=0.5\textwidth]{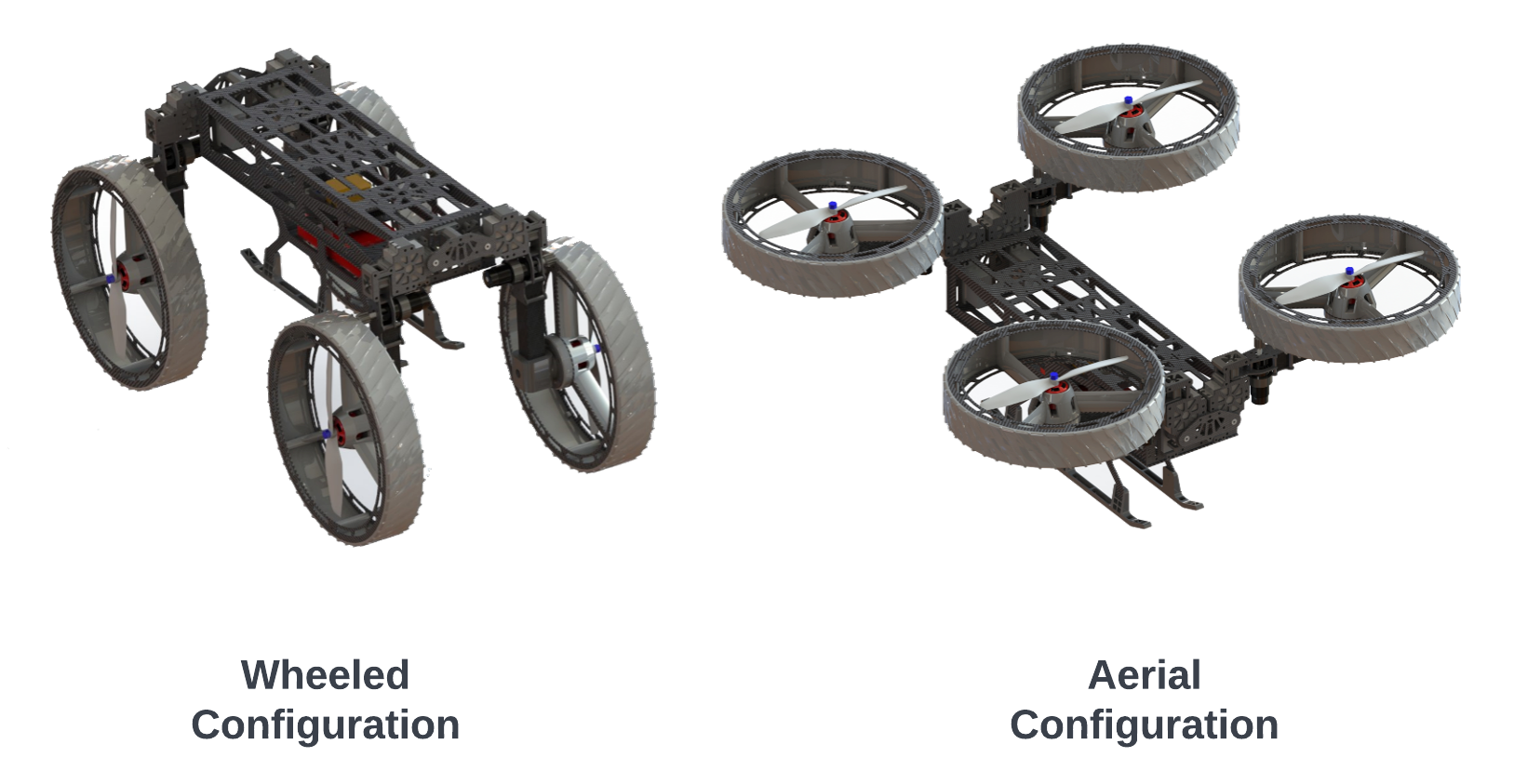}
    \caption{M4 CAD model \cite{ben}} 
    \label{fig::cad_config}
\end{figure}

The process of generating the URDF (Unified Robot Description Format) for the simulated M4 robot proved to be a time-intensive endeavor. This phase necessitated meticulous attention to detail, ensuring accurate nomenclature and alignment of joints with their corresponding joint types. Notably, the initial iteration of the URDF yielded a dense mesh file, adversely impacting simulation performance. As a resolution, a judicious transition from intricate mesh files to simplified counterparts was undertaken using Blender, executed with precision to maintain the original geometric integrity.

Post URDF development, the implementation of actuation was executed in collaboration with the ROS-control plugin, harmonizing seamlessly with Gazebo's simulation environment. The architectural underpinnings encompass three distinct actuations mirroring those of the physical counterpart: morphing transformation, flight mode, and wheel mode control. The execution of a morphing transformation, involving the conversion of M4's locomotion from wheel-driven to aerial mobility, warrants particular attention. Here, all leg joints of the robot were subjected to effort control, facilitating the seamless transition into a drone-like configuration. The core methodology governing joint control is rooted in effort modulation, which demands the meticulous calibration of PID parameters to ensure a smooth and coherent transformation. Given the concurrent manipulation of four joints during morphing, an iterative PID tuning approach was undertaken, guided by empirical trial and error.

\begin{figure}[H]
    \centering
    \includegraphics[width=0.5\textwidth]{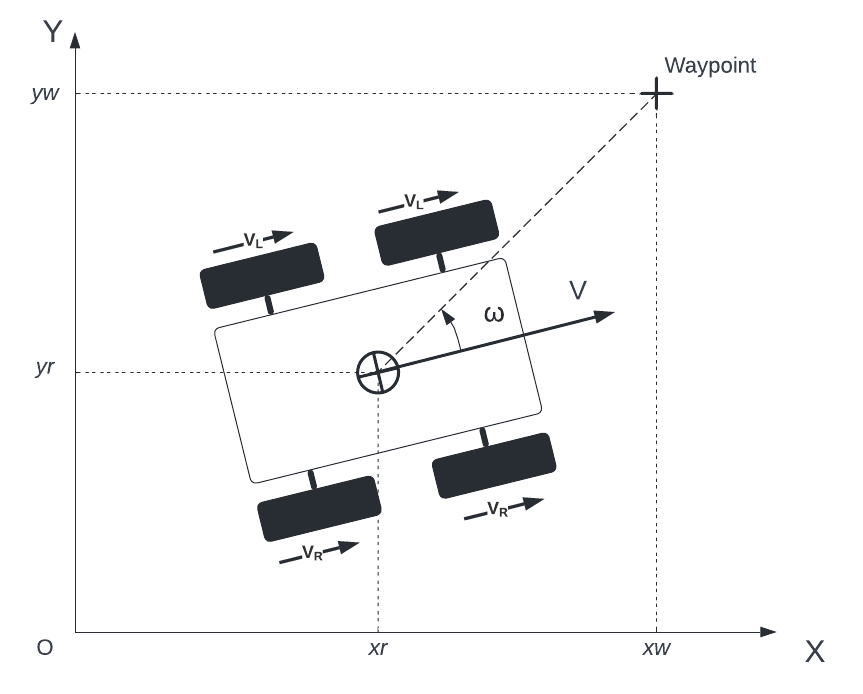}
    \caption{Representation of the differential driving control \cite{lynch2017modern}} 
    \label{fig::drive_locomotion}
\end{figure}

For the wheel mode, congruence with the physical robot was established through the integration of an identical differential drive controller. This controller effectuates directional changes by inducing variations in the spinning velocities of the left and right wheels. As a consequence, discrete motor commands, denoted as $u_L$ and $u_R$, are generated, adhering to the principles of the unicycle model. These commands are inherently contingent upon desired angular velocity ($\omega$) and forward velocity ($V$), intricately orchestrated through the 'cmd\_vel' interface. The efficacy of this locomotion control mechanism is substantiated by its execution at a frequency of 50Hz, meticulously orchestrated via the Gazebo ROS plugin, specifically the 'DiffDriveController'. Initial feasibility assessment transpired through the utilization of the RQT wheel controller, serving as a conduit for command issuance to the simulated robot. The embodiment of this locomotion schema is visually represented in Figure \ref{fig::drive_locomotion}, encapsulating the unicycle model's application to the M4 robotic system.

\begin{figure}[H]
\centering
\includegraphics[width= 1\textwidth]{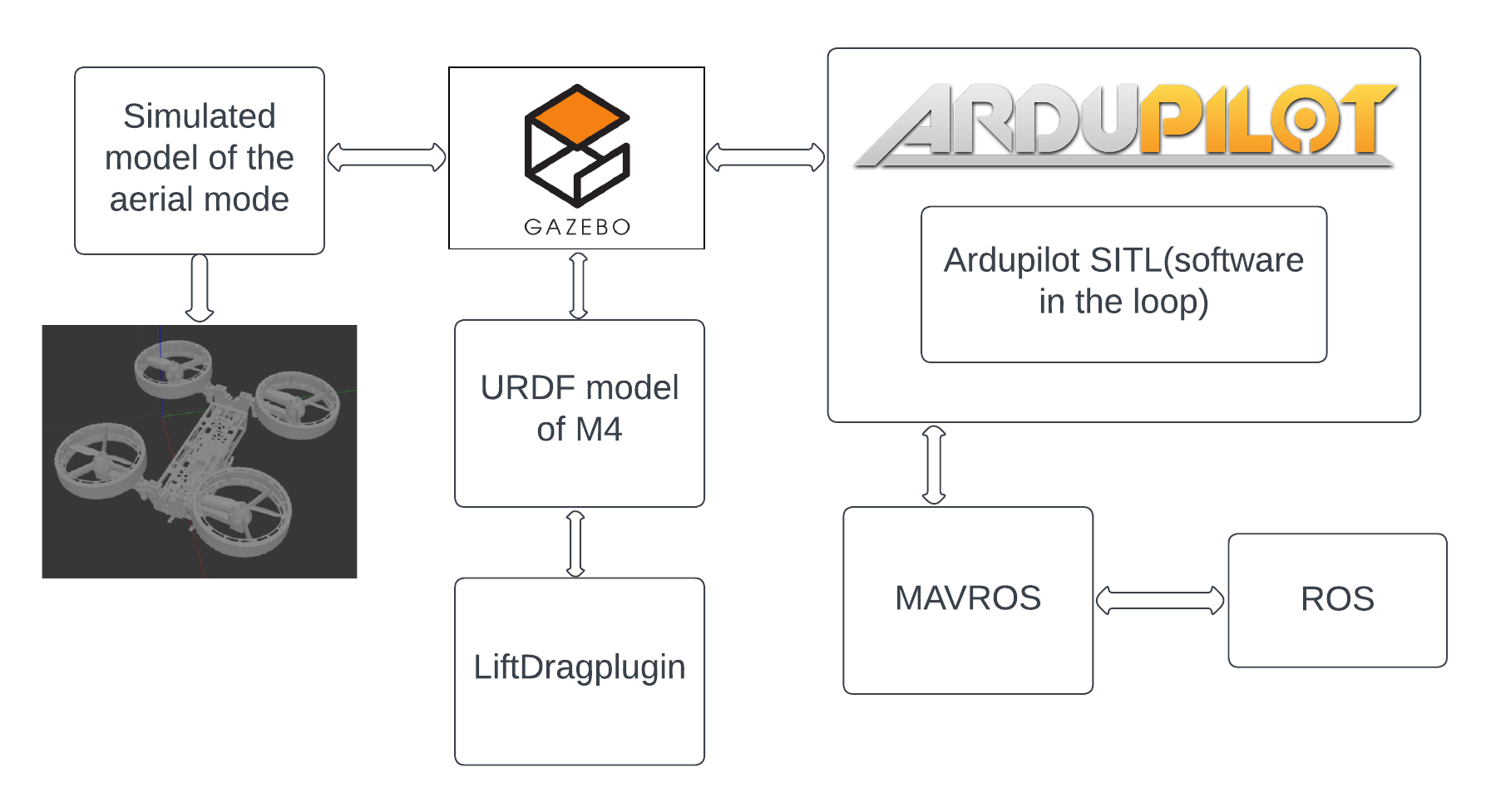}
\caption{Flight control for the model in Gazebo}
\label{fig::Ardupilot_flow}
\end{figure}

In the context of aerial mode within Gazebo, there lacked a readily available plugin for general drone control. The starting point was the hector quadcopter controls \cite{meyer2012comprehensive}, yet a transition to the M4 model necessitated changes due to weight and shape variations, rendering the original controller ineffective. Further investigation led to the realization that the real robot employed ardupilot controllers. To address this, integration between the ardupilot controller \cite{mavlink} \cite{arducontrol} and Gazebo was established using ardupilot SITL (Software In The Loop), depicted in Figure \ref{fig::Ardupilot_flow}. The functional aspects of the arducopter URDF were incorporated into the M4 URDF, and ardupilot settings were configured to match the weight changes. Subsequent testing occurred via Qgroundcontrol.

The simulation model achieved parity with the physical system in both aerial and wheel modes, particularly in terms of control. A fusion with MAVROS, a MAVLink \cite{mavlink}  extendable communication node for ROS, facilitated comprehensive robot control through a ROS architecture.

\begin{figure}[H]
\centering
\includegraphics[width=1\textwidth]{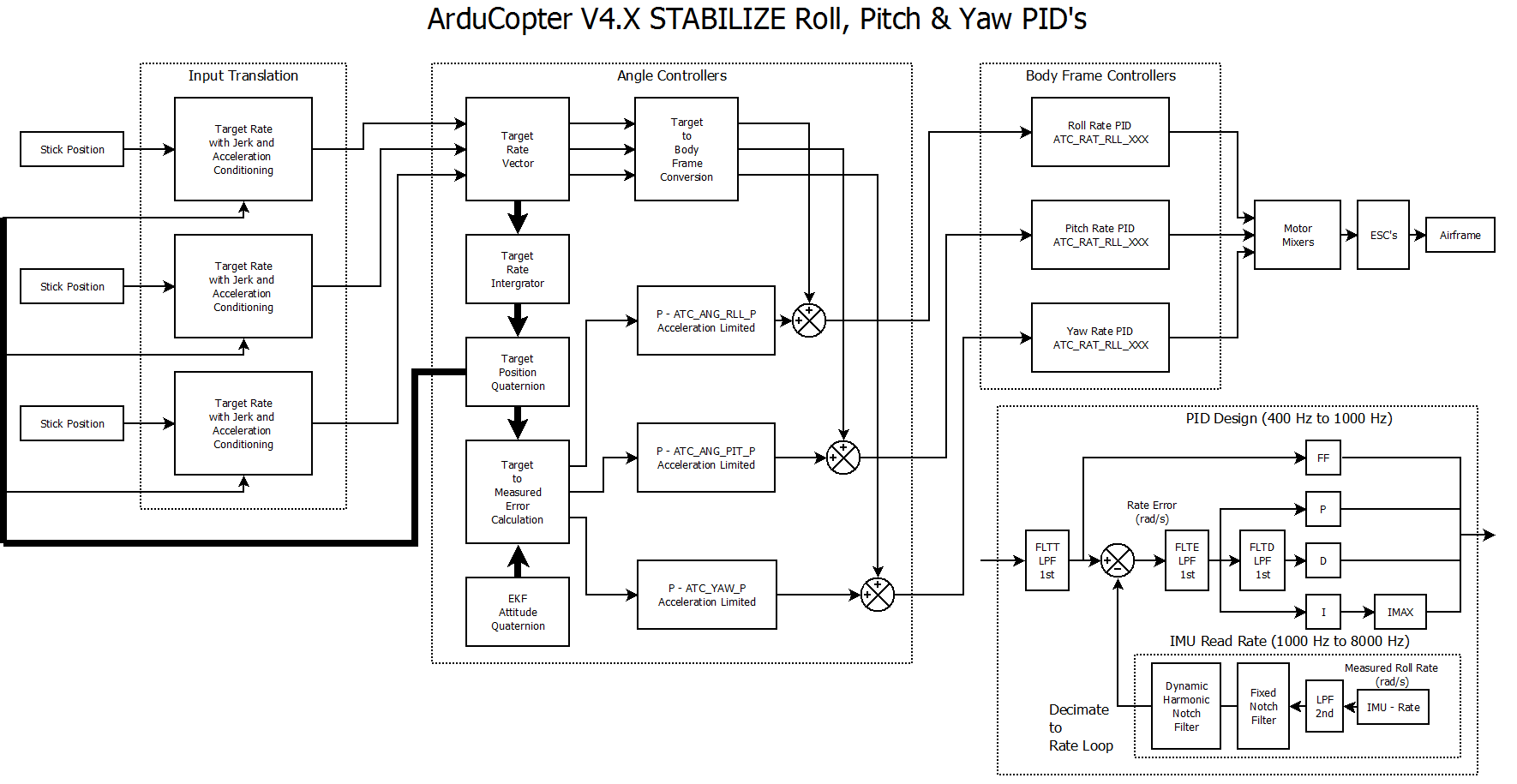}
\caption[Arducopter Altitude Control]{Arducopter Attitude Control.\footnotemark}
\label{fig::attitude_control}
\end{figure}

The ardupilot SITL \cite{ardupilot} served as middleware, receiving inputs from Gazebo's IMU and position simulations. These inputs were subsequently linked to arducopter control \cite{ardupilot} mechanisms, enabling altitude control (Figure \ref{fig::attitude_control}). The guided mode was employed to furnish the robot with waypoints for assessment.

\footnotetext{\url{https://ardupilot.org/dev/docs/apmcopter-programming-attitude-control-2.html}}

Post-integration of distinct controls into the simulation model, a RealSense camera was incorporated as a perception sensor, mimicking the real robot's setup. Utilizing an existing Gazebo plugin for RealSense, the necessary adjustments to the URDF and topic were made. Additional insight revealed the requirement for an extra sensor during flight; a bottom-mounted sensor was included beneath the robot's main body, preceding its landing mount. This augmentation facilitated environment mapping within the Gazebo.

In terms of localization, Gazebo's ground truth was employed to provide the real ground location of the robot. To emulate real-world conditions, some noise was intentionally added to this data. This noisy localization information was then published to serve as the robot's odometry with respect to the world frame.

Custom environments, created using Blender for uneven terrain and Gazebo for obstacle insertion, were utilized to scrutinize the robot's capabilities in multimodal locomotion and its autonomous pipeline.

\section{Perception and Navigation Pipeline}
\label{sec::pipeline}

The perception and navigation pipeline forms the backbone of autonomous robotic systems, enabling them to understand their surroundings and make informed decisions. This chapter presents an overview of our pipeline, highlighting the integration of sensor technologies, perception algorithms, and intelligent planning strategies. Through these components, our methodology equips the robot to efficiently perceive its environment and navigate the surroundings, ensuring safe and effective interactions within real-world scenarios.

\begin{figure}[H]
\centering
\includegraphics[width=1\textwidth]{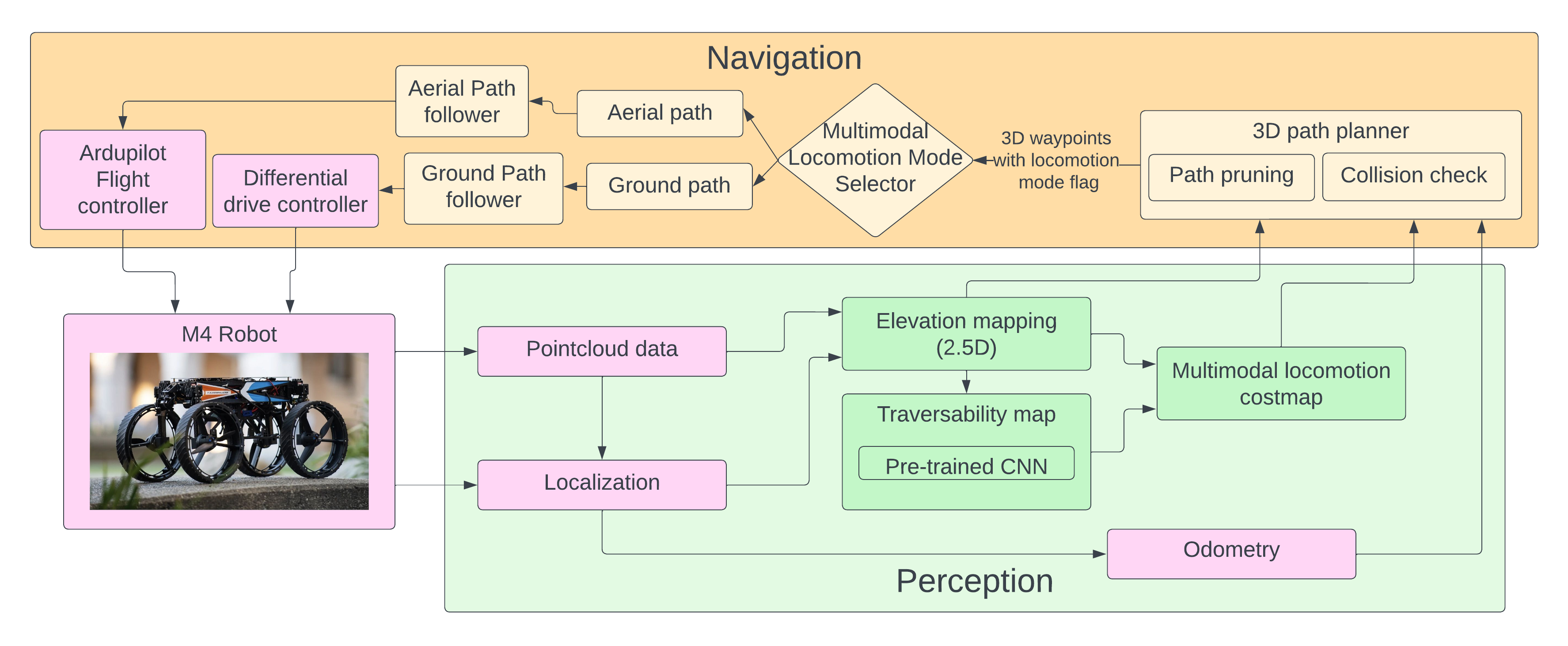}
\caption{Whole Autonomous pipeline for M4 Robot}
\label{fig::main_pipeline}
\end{figure}

above figure \ref{fig::main_pipeline} shows the overview of the system and the whole pipeline design, we will go deep into each aspect of the above and explain each part.
take note that the pipeline above is designed for both the real world and simulation.
for the real-world the localization and pointcloud data is taken from realsense camera and Rtabmap\cite{RTABMAP} whereas in the gazebo simulation environment to keep things simple we use the ground truth of the gazebo as the localization of the robot and pointcloud data is taken from simulated realsense camera in the gazebo.

\begin{figure}[H]
    \centering
    \includegraphics[width=1\textwidth]{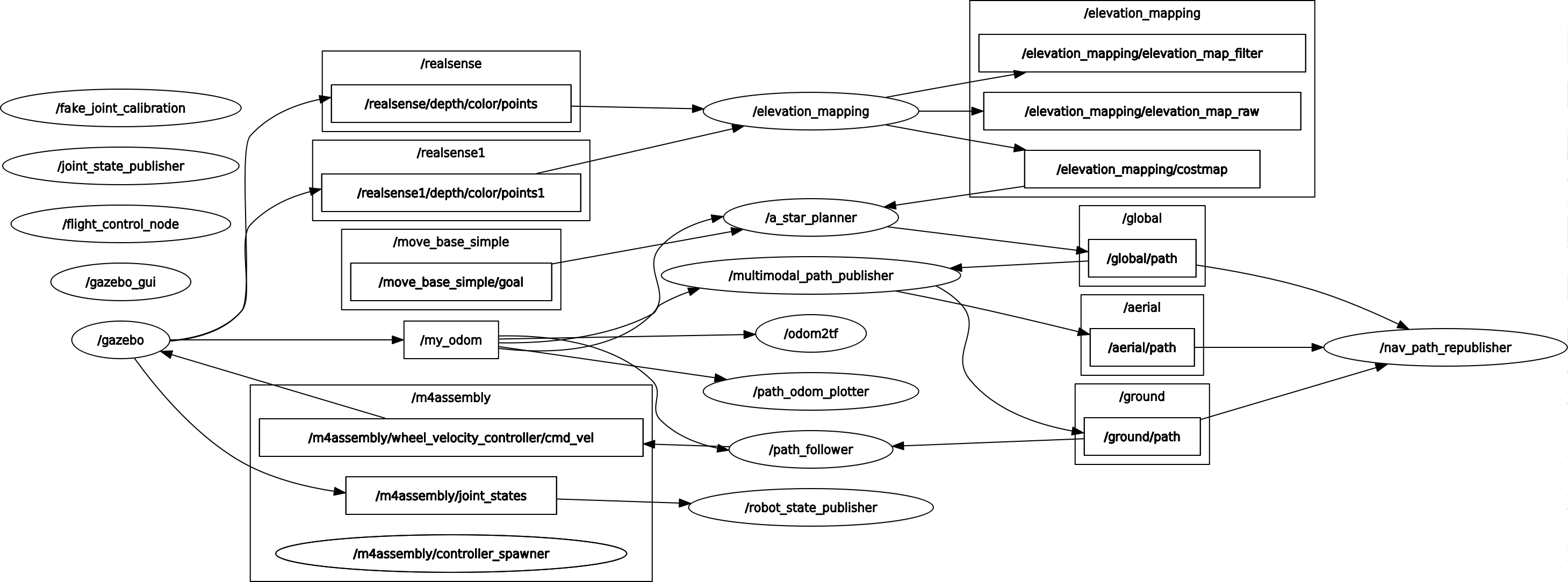}
    \caption{Rosnode image, the whole pipeline when running in ROS noetic, from the left side you can see the gazebo and it's giving out realsense pointcloud data and ardupilot nodes, then from there the elevation mapping and path planning nodes are connected which then goes back to the simulation} 
    \label{fig::rosnode}
\end{figure}

\subsection{Elevation mapping}
\label{subsec::elevation mapping}

Contemporary approaches in Multimodal locomotion robots often advocate for the use of full three-dimensional (3D) mapping techniques to facilitate the perception and decision-making processes. However, after careful consideration of the specific requirements and operational constraints of the Multi-Modal Mobility Morphobot (M4) Robot, the adoption of a 2.5D mapping approach emerges as the preferred choice.

This decision is rooted in a strategic balance that marries the advantages of a richer spatial understanding with the imperative of real-time adaptability and energy-efficient performance. Embracing a 2.5D mapping approach offers the following benefits:

\begin{itemize}
\item Adapt to Varied Terrain: The robot's capability to navigate diverse terrains is a critical factor. 2.5D maps capture elevation changes and basic terrain features, allowing the robot to plan trajectories that account for steps, slopes, and other ground variations.

\item Enhanced Depth Perception: The 2.5D mapping approach provides depth information that is crucial for the Multi-Modal Mobility Morphobot (M4) operating in complex environments. Unlike traditional 2D maps, which lack vertical information, and full 3D maps, which might be computationally expensive, 2.5D maps offer a balanced representation that enables the robot to understand and navigate through both horizontal and vertical obstacles.

\item Maintain Real-time Responsiveness: The Multi-Modal Mobility Morphobot (M4) necessitates swift and agile responses to its environment. 2.5D mapping strikes a balance between depth perception and computational efficiency, ensuring the robot can make informed decisions without compromising on speed.

\item Computational Efficiency: Processing complete 3D maps demands significant computational resources. 2.5D mapping requires less processing power compared to full 3D mapping while providing substantial depth cues. This efficiency is crucial for the Multi-Modal Mobility Morphobot (M4), which needs to optimize computational resources for both flight control and ground navigation.

\item Human Interaction: The 2.5D map is intuitively interpretable for human operators, facilitating effective communication and collaboration. Operators can quickly assess the environment and provide guidance to the robot during complex missions. This advantage is particularly valuable in scenarios involving search and rescue, surveillance, or exploration.
\end{itemize}
In conclusion, the strategic adoption of the 2.5D mapping approach aligns seamlessly with the Multi-Modal Mobility Morphobot (M4) robot's mission objectives, ensuring it remains efficient, responsive, and adaptable in a range of challenging environments.

\subsubsection{Package for 2.5D mapping}
\label{subsec::package2.5Dmapping}

The "Elevation Mapping Cupy" \cite{mikielevation2022}package, crafted by the esteemed Robotic Systems Lab at ETH Zürich, takes inspiration from ANYbotics grid map \cite{gridmap} and Robot-Centric Elevation Mapping methods \cite{elevation1,elevation2}. This package is pivotal for our Multi-Modal Mobility Morphobot (M4) robot's 2.5D elevation mapping. We chose it based on practicality and innovation. It's open-source, letting us customize it to our project's needs. The package's GPU acceleration boosts our computing power, making us faster and more responsive real-time. Moreover, it's cutting-edge, fitting right in with our advanced goals. Its clever algorithms match our need for sharp perception. Plus, it sets the stage for future integration with deep learning techniques for our robot.

\begin{figure}[H]
\centering
\includegraphics[width=0.75\textwidth]{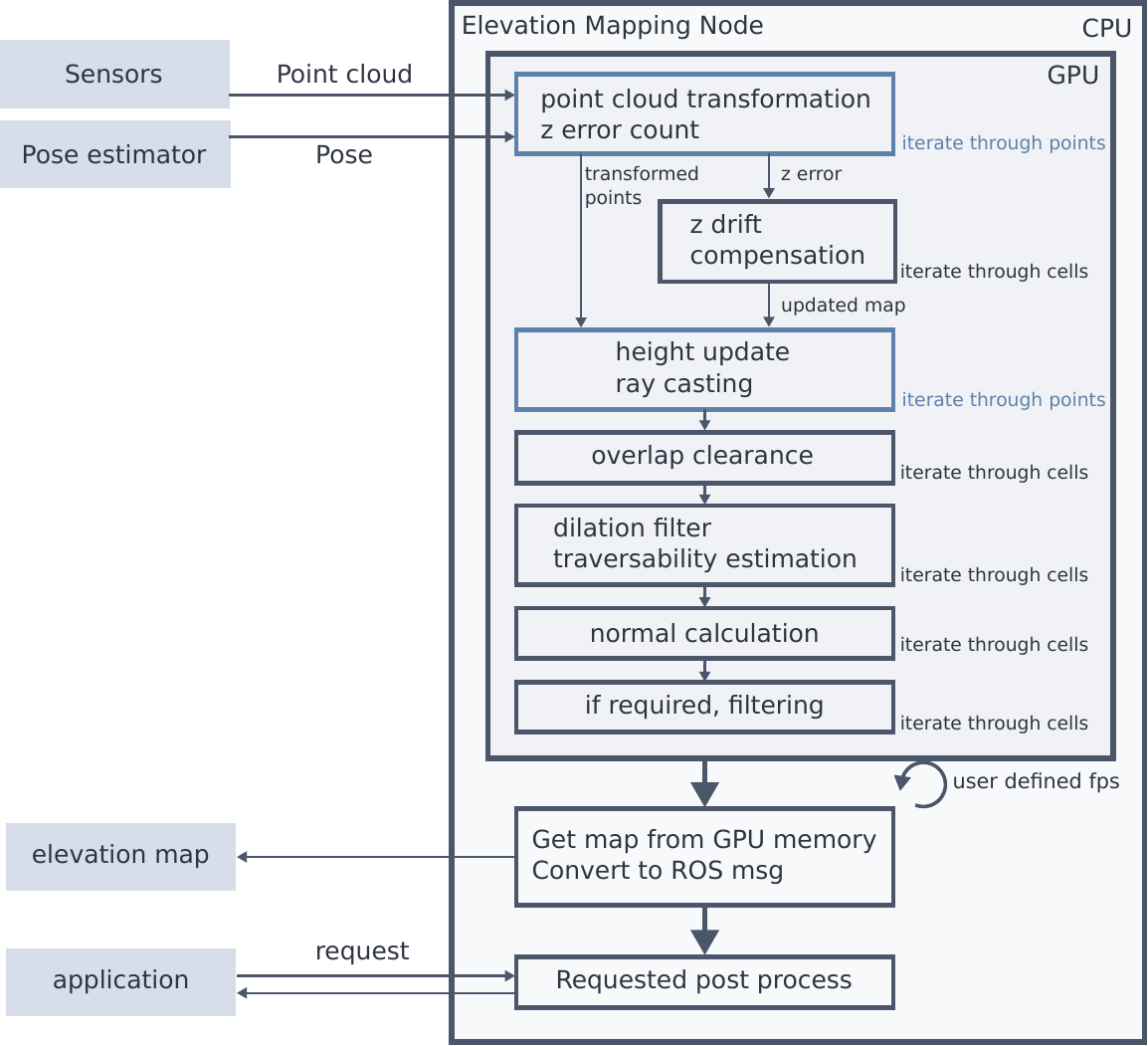}
\caption{Overview of processing. The point cloud data from the sensor
and the pose information from the pose estimator are inputs to the GPU.
After transforming the point cloud and calculating the z-drift errors, the
current map estimation is adjusted to match the latest sensor measurement.
Then the main height map update and ray casting are performed. As the
map is updated, various filters are applied. The map data is transferred to CPU memory with a user-defined frequency to publish as a ROS message
to reduce unnecessary data transmission.\cite{mikielevation2022}} 
\label{fig::cupy}
\end{figure}

Figure \ref{fig::cupy} provides an insightful representation of the elevation mapping creation process facilitated by the "Elevation Mapping Cupy" package \cite{mikielevation2022}, and its operation unfolds as follows:

The process commences with input in the form of point cloud data derived from depth sensors, in addition to the robot's estimated pose, acquired via Simultaneous Localization and Mapping (SLAM) based pose estimation or an odometry system. These inputs collectively steer the elevation mapping process, achieved exclusively through the capabilities of the "Elevation Mapping Cupy" package \cite{mikielevation2022}.

\begin{figure}[H]
\centering
\includegraphics[width=0.5\textwidth]{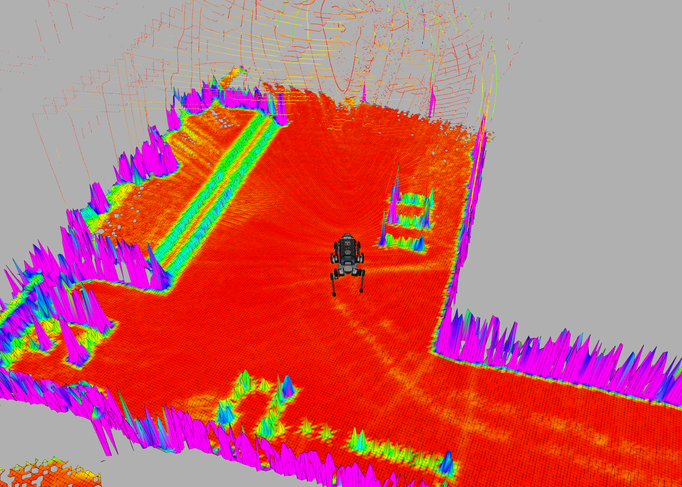}
\caption{sample elevation mapping\cite{mikielevation2022}} 
\label{fig::elevationmapping}
\end{figure}

To begin, the data is seamlessly transferred to the GPU memory, capitalizing on the package's GPU acceleration for swift and efficient computation. The ensuing step involves the transformation of points into the user-specified map frame using the ROS tf library, thereby aligning the data for coherent representation.

A noteworthy aspect of the process is the calculation and storage of height drift error, a critical metric instrumental for subsequent phases. This height error data is used to effectuate a synchronization between the map and the most recent sensor measurements, effectively fine-tuning the map's alignment.

Once the map is optimized with the latest measurements, the software embarks on the iterative journey of height update, where each point measurement contributes to the ongoing refinement of the map's elevation data. Concurrently, the package employs ray-casting to eliminate objects that may have penetrated the map, enhancing the map's accuracy.

Further advancing map accuracy, a series of essential operations are executed, encompassing overlap clearance, estimations of traversability, calculations of normals, and potential filtering to enhance the map's quality.

Ultimately, the culmination of these processes leads to the publication of the elevation map at a frequency defined by the user. The map data transitions from the GPU to CPU memory, forming a GridMap message as per  \cite{gridmap}, a strategic move aimed at optimizing data transfer and resource utilization.

It is important to note that the described processes are seamlessly orchestrated by the "Elevation Mapping Cupy" package itself \cite{mikielevation2022}, with no direct involvement from our end. This package demonstrates its prowess by automating the height updates within each cell, utilizing a Kalman filter formulation for this purpose, as previously established \cite{elevation1,elevation2}.

the package takes in depth pointcloud at 30 Hz and then gives out elevation mapping topic at 5 Hz.This speed is the fastest the map can be made and all other plugins can work below or with this speed for publishing.
\subsection{Traversability map using deep learning}
\label{subsec::traversability}
Traversability for a robot means whether or not that robot can go over the given environment and how easy or difficult it is for the robot. the traversability is widely used in legged mobile robots as they can traverse some uneven terrains that cannot be detected by 2D occupancy mapping means. 
Deriving traversability from the map is very important for the robots, there are many ways to find traversability, we started by checking for the classical approach to traversability by using a classical approach which uses slope, and step filters and get the traversability out by \cite{traversability_classical}.

Performing elevation mapping on a GPU offers a significant advantage by capitalizing on the data residing in memory. This eliminates the processing overhead associated with data transfers between the CPU and GPU. An efficient terrain analysis utilizing neural networks, specifically a Convolutional Neural Network (CNN) model trained to predict traversability values for robot navigation, becomes feasible in this context \cite{wellhausen2021rough}. The CNN, implemented using PyTorch \cite{pytorch}, is lightweight enough to execute seamlessly at the map's full update rate.\cite{mikielevation2022}.
the CNN currently used is by \cite{wellhausen2021rough} which is originally based on \cite{main_traversability}

\begin{figure}[H]
\centering
\includegraphics[width=1\textwidth]{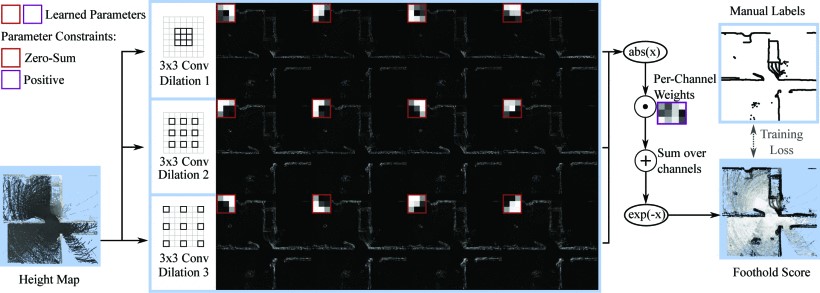}
\caption{A shallow convolutional network was trained to forecast foothold scores from elevation maps, relying on a limited dataset of 20 manually annotated training instances. The visualization of learned convolutional weight parameters across various dilations, alongside their subsequent activation on a sample input height map, is depicted at the center. The application involves multiplying the absolute value of each filter channel by a learned weight. Ultimately, the foothold score, integral for gauging stability, is derived from the negative exponential of the collective channel summation. This scoring mechanism is honed through supervised learning using meticulously annotated reference examples. (Adapted from\cite{wellhausen2021rough}} 
\label{fig::cnn}
\end{figure}

\begin{figure}[H]
\centering
\includegraphics[width=1\textwidth]{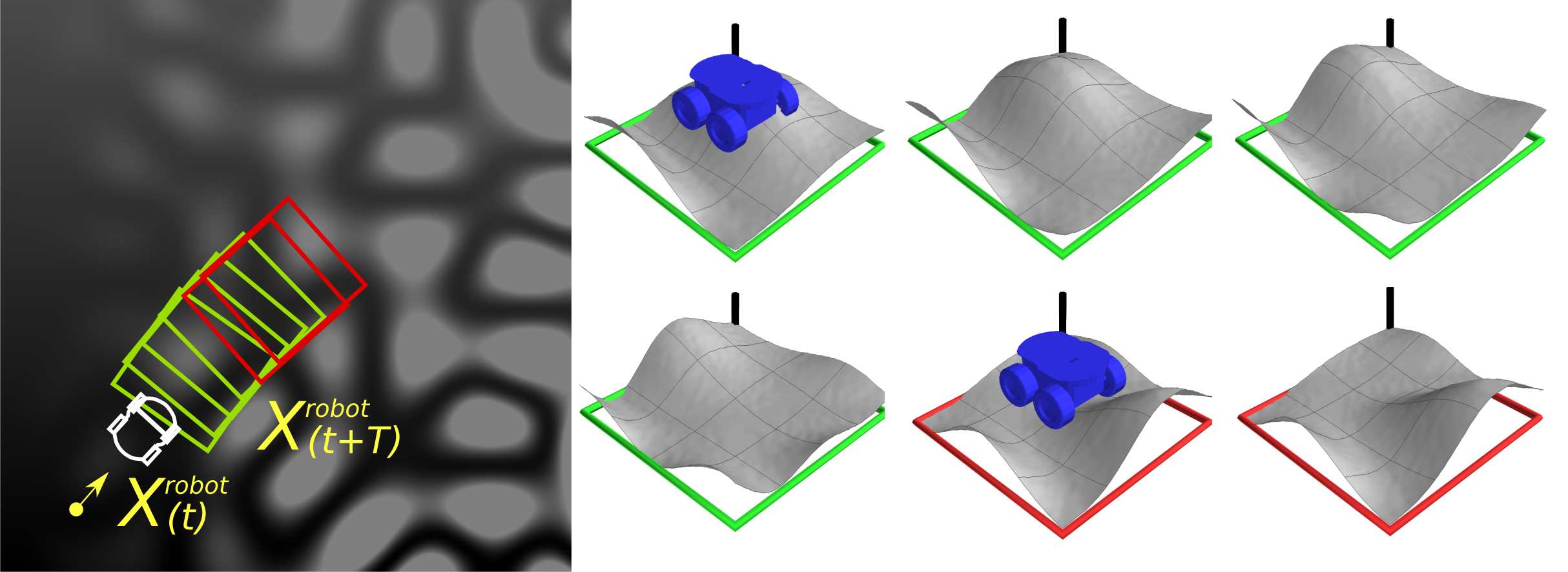}
\caption{Left: trajectory extracted from a training dataset; robot silhouette
and yellow arrow indicate the initial pose. Right: visualization of some of
the resulting traversable (green) and non-traversable (red) patches\cite{main_traversability}} 
\label{fig::traversedetect}
\end{figure}

The above network worked great as initial testing showed good results even for mobile robots as it was trained on manual data it gave nice traversability at real-time, due to that reason at this point for the scope of the thesis training a new network was redundant, so we decided to move ahead with the same traversability CNN network to test our pipeline.we did notice that as the dataset was for foothold score of legged robot some of the high ceiling obstacles had large non-traversable areas suitable for legged robots but excessive for mobile robots so for future works, To recifity that I have created the gazebo simulation pipline as per \cite{main_traversability} for getting multimodal traversability out for wheeled ,walking and segway locomotion traversability and cost to the robot as form fig \ref{fig::traversedetect} the traversability as for the above networks also was trained via gazebo simulation where they created different heightmaps and ran the robot through them and as they moved they saw if the robot could travel or not or will it require more effort and they too the photos of the robots trajectory as fig \ref{fig::traversedetect} and determined the traversability for a patch.after creating dataset from multiple runs of gazebo with different environment they trained a CNN network to determine traversability with the input of heighmap or elevation map which is the same concept applied to M4 very nicely as in gazebo we can collect wheel traversability with its cost and also collect segway \ref{fig:morph} traversability while collecting its energy consumption. to that end for the whole testing we used \ref{fig::cnn} but i have replaced the robot in the gazebo and have intergrated the M4 robot but the CNN sides need work to optimze it to work for real-time operation which is why it was excluded from this thesis scope.

The initial network exhibited remarkable performance during preliminary tests, particularly for mobile robots. Benefiting from its training on manual data, the network demonstrated commendable real-time traversability results. Given these achievements and the scope of the thesis, developing a new network was considered redundant. Consequently, the decision was made to proceed with the existing Convolutional Neural Network (CNN) for traversability testing within the proposed pipeline.
\begin{figure}[H]
\centering
\includegraphics[width=0.8\textwidth]{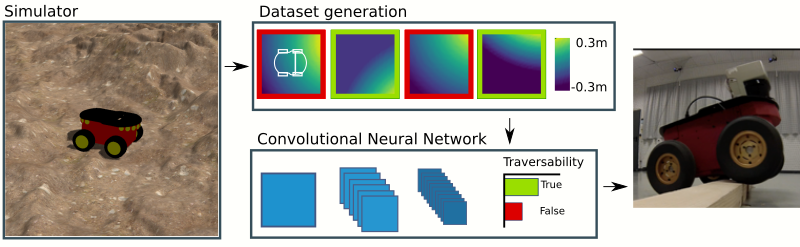}
\caption{Pipeline for training the traversability for a custom robot\cite{main_traversability}} 
\label{fig::traversepipeline}
\end{figure}
The dataset, originally intended for evaluating legged robot foothold scores, presented challenges when applied to mobile robots navigating high-ceiling obstacles. These obstacles featured extensive non-traversable areas suitable for legged robots but excessive for mobile robots. To address this limitation, a Gazebo simulation pipeline was developed following the methodology outlined in \cite{main_traversability}. This pipeline generated multimodal traversability assessments for various locomotion modes, including wheeled, walking, and segway. The simulation involved generating diverse heightmaps, simulating robot traversal, and capturing trajectory images (Figure \ref{fig::traversedetect}). The resulting dataset was used to train a CNN network to determine traversability based on elevation maps. This approach was found to synergize effectively with the M4 robot, enabling the collection of wheel traversability data with associated costs, as well as segway traversability and energy consumption assessments (Figure \ref{fig:morph}). However, while testing the proposed pipeline, the CNN architecture had 7 layers not feasible for operating on jetson-level hardware in real-time and required further optimization to achieve real-time operation, a task that fell outside the scope of this thesis.

\subsection{Multimodal Locomotion Costmap and autonomous mode selection}
\label{subsec::costmap}

\begin{figure}[H]
\centering
\includegraphics[width=1\textwidth]{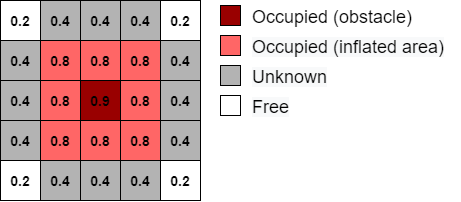}
\caption{sample costmap from occupancy grid\cite{costmap}} 
\label{fig::simple_costmap}
\end{figure}

Traditional occupancy costmaps\ref{fig::simple_costmap}, although widely used for navigation, may fall short in accommodating the diverse locomotion capabilities of multimodal robots. These costmaps primarily focus on representing obstacles and free spaces as binary values, often relying on sensor data to mark occupied and unoccupied cells. While effective for ground-based wheeled robots, they can pose limitations for multimodal robots that can switch between different locomotion modes, such as walking, flying, or climbing.

Multimodal robots require more nuanced costmaps that consider not only obstacles but also factors specific to each locomotion mode. For instance, an obstacle that is impassable for a ground robot might be traversable for an aerial drone. Additionally, variations in terrain traversability, energy efficiency, and stability must be accounted for differently in different modes. Traditional occupancy costmaps do not inherently provide the granularity needed to differentiate between these diverse requirements.

The "Multimodal Locomotion Costmap Generation Algorithm"\ref{alg::costmap} is designed to facilitate efficient robotic navigation across varying terrain conditions and locomotion modes. Given input data including elevation information from the $elevation\_map$ and traversability details from the $traversability\_map$, the algorithm generates a costmap layer that guides the robot's movement.

The process begins with initializing a costmap layer and subsequently updating it based on the traversability of each cell. When traversability is below a threshold of 0.5, indicating challenging terrain or untraversable terrain, the algorithm enhances the costmap values by combining the elevation data from the $elevation\_map$ with a small multiplier and tags the area with flying locomotion mode. This approach allows the costmap to account for uneven terrain while flying to inflate the cost of higher elevation slopes or beams.

\begin{algorithm}[H]
\caption{Multimodal locomotion Costmap Generation Algorithm}\label{alg::costmap}
\KwData{$elevation\_map$,$traversability\_map$,$inflation\_radius$,
$L{e} -locmotion\_modes\_energy\_ratio= 60$ }
\KwResult{Generated Multimodal locomotion costmap layer}

\BlankLine
\textbf{Initialization:}

Initialize $costmap\_layer$ as a zero matrix\;
Extract $traversability$ and $elevation\_map$;
\BlankLine
\textbf{Update Costmap Layer:}

\ForEach{cell $(i, j)$ in $costmap\_layer$}{
    \eIf{$traversability[i, j] < 0.5$}{
        Set $costmap\_layer[i, j] = L{e} + elevation\_map[i,j]*0.01$\;
        \CommentSty{//adding elevation value as a way to inflate uneven heights so that path is optimal}
    }{
        Set $costmap\_layer[i, j] = 1 - traversability[i, j]$\;
        \CommentSty{//Adding different traversability for mobile energy efficiency}
    }
}
\BlankLine
\textbf{Inflation:}

Identify obstacle cells as those with values above 60\;

\ForEach{obstacle cell $(i, j)$}{
    $value \gets costmap\_layer[i, j]$\;
    Extract $neighborhood$ of $(i, j)$\;
    $neighborhood \gets \text{where}(neighborhood < value, value, neighborhood)$\;
    Update $costmap\_layer$ with inflated $neighborhood$\;
}

\BlankLine
\Return{$costmap\_layer$}
\end{algorithm}

Conversely, for cells with traversability above 0.5, representing more favorable terrain, the algorithm sets the costmap values to be inversely proportional to traversability. This aims to promote energy-efficient paths, encouraging the robot to preferentially choose paths that minimize energy consumption.

The algorithm also considers inflation around obstacles. Obstacle cells are identified based on their higher cost values, typically denoting aerial nodes in the environment. The algorithm then inflates the costmap values in the vicinity of these obstacles, ensuring a safety margin around them to guide the robot's path planning for the other locomotion.

It's important to note that the generated costmap is published at a rate of 1 Hz. This means that the algorithm calculates and updates the costmap layer once per second, providing the robot with real-time guidance for safe and efficient navigation through its surroundings. Overall, this algorithm plays a crucial role in enabling multimodal locomotion, enhancing both safety and energy efficiency, and providing robots with a reliable tool for navigating complex terrains.

\subsection{Efficient 3D Path Planning with Energy Considerations through 2.5D mapping}
\label{subsec::Path planner}

The main goal of path planning is to create a sequence of points that guides a robot from one position to another while staying within the environment's limits. Another important aim is to optimize this path to use less energy, which makes the robot more independent by using different ways of moving. It's also crucial to think about how much computer power is needed to run the path planning algorithm and mapping in a reasonable time. This is especially important when dealing with a 3D environment full of obstacles and trying to figure out the best ways for the robot to move in different ways.

Right now, the M4 project has all the locomotion modes implemented but as per the thesis scope we are focusing on flying and ground locomotion mode. In this work, to limit the scope of the thesis a static environment without dynamic obstacles was considered which can be further expanded by using a local planner for dynamic obstacles easily in the future.

\begin{figure}[H]
    \centering
    \includegraphics[width=0.3\textwidth]{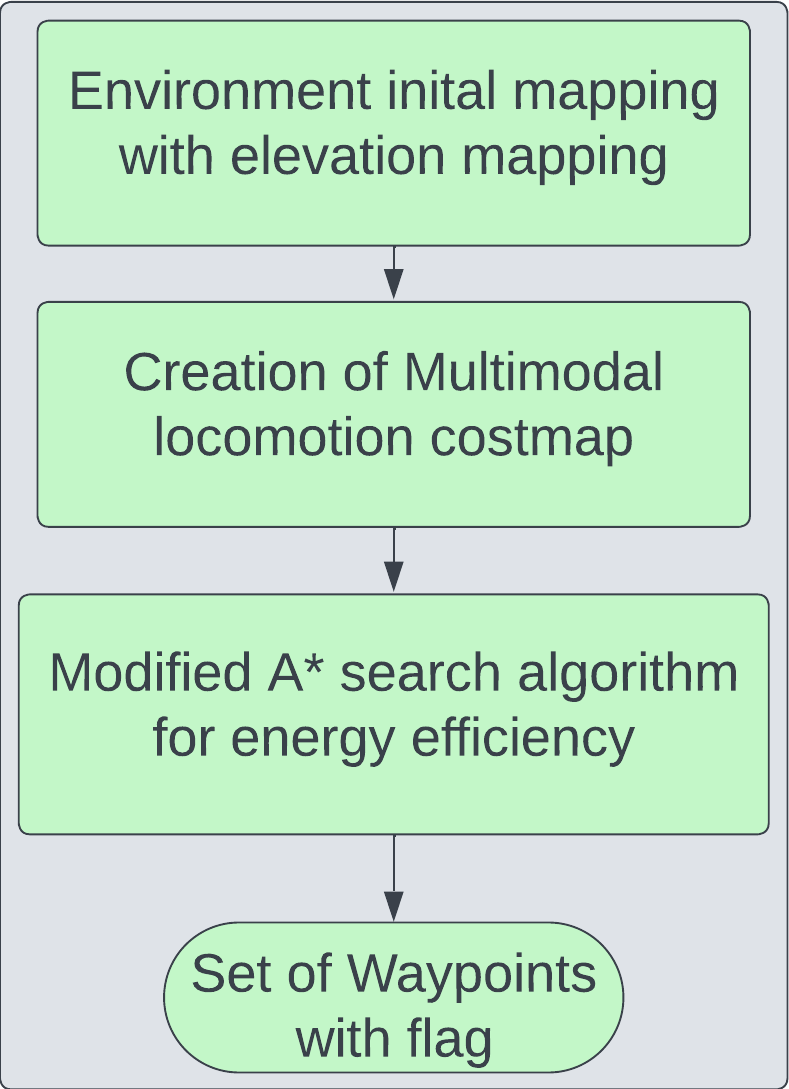}
    \caption{Main steps of the Path Planning Method.} 
    \label{fig::global_path_planning}
\end{figure}

Our path-planning process can be divided into four steps, which you can see in Figure \ref{fig::global_path_planning}. First, The environment must be mapped out by the depth camera or using known external resources of the M4 robot in both simulation and real-world scenarios. Then, we use already uniformly descertized 2.5D maps and our costmap to create an efficient 3D path with just 2.5D information which reduces the dimensionally of the navigation planner and results in faster computation. Then we do a collision check for the elevation map and give out the optimum path.

To find the optimal path in the map, the Modified A$^\star$ path search algorithm based on standard A$^\star$\cite{A_star} is used. This modified
A$^\star$ algorithm is employed to find the optimal path by using a custom heuristic function. In doing this, a new and Multimodal locomotion oriented 3D version of the  A$^\star$ for 2.5D maps is implemented. The algorithm computes the best path to each node only to visit the most promising ones based on our heuristics. The modified algorithm is as shown in Algorithm\ref{alg:3D_Astar}.

\begin{algorithm}
\caption{Modified 3D A* Path Planning for 2.5D elevation map}\label{alg:3D_Astar}
\SetAlgoNlRelativeSize{0}
\SetNlSty{}{}{:}
\SetAlgoNlRelativeSize{-1}

\SetKwProg{Algorithm}{Algorithm}{:}{}
\SetKwFunction{GetNeighbors}{GetNeighbors}
\SetKwFunction{WorldToGrid}{WorldToGrid}
\SetKwFunction{MaxElevation}{MaxElevation}
\KwIn{$s$: start position, $g$: goal position,costmap,elevation}
\KwOut{Path $p$ from $s$ to $g$}

$\text{open} \gets [(0, s)]$\,$\text{came\_from} \gets \{\}$\;
$g\_s \gets \{s: 0\}$\,$f\_s \gets \{s: \text{h}(s, g)\}$\;

\While{$\text{open}$ is not empty}{
    $c \gets \text{priority(open)}[1]$\;
    
    \If{$\text{h}(c, g) \leq \text{threshold}$}{
        $p \gets [x,y,z,flag]$\;
        \KwRet $p$\;
    }
    
    \For{$n$ in $\text{get\_n}(c)$}{
        $(x, y) \gets \text{w\_to\_g}(n)$\,$(x0, y0) \gets \text{w\_to\_g}(c)$\;
        
        $c\_n \gets \text{costmap}[x, y]$\,$e\_n \gets \text{elevation}[x, y]$\,$e\_c \gets \text{elevation}[x0, y0]$\;
        $\Delta e \gets \text{round}\left(\frac{\lvert e\_c - e\_n \rvert}{\text{res}}\right)$\;
        
        \If{$c\_n =< 60$}{
            $\text{flag} \gets \text{l\_f}$\;
            $z \gets \text{max\_e}(x, y) + \text{cl}$\;
            $g\_n \gets g\_s[s] + \text{h}(c, n) + c\_n \times \Delta e + cle_{a} + c\_n$\;
        }
        \Else{
            $\text{flag} \gets \text{l\_w}$\;
            $z \gets e\_n$\;
            $g\_n \gets g\_s[s] + \text{h}(c, n) + c\_n$\;
        }
        
        \If{$n$ not in $g\_s$ \textbf{or} $g\_n < g\_s[n]$}{
            $\text{came\_from}[\text{tup}(n)] \gets c, z, \text{flag}$\;
            $g\_s[n] \gets g\_n$\;
            $f\_s[n] \gets g\_n + \text{h}(n, g)$\;
            $ Priority \gets [f\_s[n],\text{tup}(n)]$\;
        }
    }
}
\end{algorithm}

The provided algorithm describes a modified version of the 3D A* path planning algorithm tailored for navigating through a 2.5D elevation map. The algorithm takes into account both the horizontal grid positions and vertical elevation values to efficiently find a path from a given start position ($s$) to a goal position ($g$). The input to the algorithm includes the start and goal positions, a costmap representing traversal costs for each grid cell and from that inferring their locomotion mode possibility, and an elevation map providing elevation values for each cell. The output is a path ($p$) consisting of waypoints with information about their coordinates ($x$, $y$, $z$) and a flag indicating whether the waypoint is an Aerial(\text{l\_{f}}) or wheeled(\text{l\_{w}}).\ref{alg:3D_Astar}

Symbols in the algorithms and following equations mean:
\begin{align*}
    \text{h}(a, b) & : \text{Heuristic function between nodes } a \text{ and } b \\
    \text{tup}(x) & : \text{Convert } x \text{ to a tuple representation} \\
    \text{get\_n}(c) & : \text{Get neighbors of node } c \\
    \text{w\_to\_g}(p) & : \text{Convert world coordinates } p \text{ to grid coordinates} \\
    \text{costmap}[x, y] & : \text{Cost of traversing grid cell at } (x, y) \\
    \text{elevation}[x, y] & : \text{Elevation value at grid cell } (x, y) \\
    \text{max\_e}(x, y) & : \text{Maximum elevation at neighboring cells of } (x, y) \\
    \text{cl} & : \text{Clearance value} ,\text{res} : \text{Resolution of grid} \\
\end{align*}

The Heuristic Function $h(a, b)$ for getting the distance between the start and current points is as below which uses $x$, $y$ of the grid to calculate the initial cost which is then added by the elevation cost.
\begin{equation}\label{eq:g(n)}
\text{h}(a, b) = \sqrt{(b_x - a_x)^2 + (b_y - a_y)^2}
\end{equation}
The whole time the global coordinate system of the system is wrt the world frame but the gridmap\cite{gridmap} library has grid data storage with a different axis which is at the center of the grid map, to convert it we basically divide the world co-ordinates with resolution Transformation where we know the height and width of the grid map so we figure out the world to grid position and the same way we find the grid to the world by using the inverse of the equations as shown below.
\begin{equation}
\text{pos\_x} = \left( \frac{\text{height}}{2} + \text{origin}[0] + \text{pos}[0] \right)
\end{equation}
\begin{equation}
\text{pos\_y} = \left( \frac{\text{width}}{2} + \text{origin}[1] + \text{pos}[1] \right)
\end{equation}
\begin{equation}
\text{WorldToGrid}(pos) = \left(\frac{pos_x}{\text{res}}, \frac{pos_y}{\text{res}}\right)
\end{equation}

the max elevation determines the z height so that the robot measures the whole area where it occupies the space and takes the max value to avoid a collision. 

\begin{equation}
\text{MaxElevation}(x, y) = \max
\begin{pmatrix}
\text{elevation}[x, y] & \text{elevation}[x+1, y] \\
\text{elevation}[x, y+1] & \text{elevation}[x+1, y+1]
\end{pmatrix}
\end{equation}

Finally, when each time the algorithm explores the n-th node, it calculates the minimum cost $f(n)$ necessary to reach the goal by passing through it using the following formula:
\begin{equation}
    f(n) = g(n) + h(n),
\end{equation} 
\noindent where $g(n)$ is the real cost from the start to the n-th node, computed based on \eqref{eq:g(n)}, and $h(n)$ denotes the heuristic cost to the goal. The heuristic cost $h(n)$ is calculated by Euclidean distance between the two points in x,y,for each node point the $g(n)$ value is used by the following the cost :

For Aerial ($l_f$) Nodes:
\begin{equation}
g_{n_{\text{aerial}}} = g_{s}[s] + h(c, n) + c_{n} \cdot \Delta e +cle_{a} + c_{n}
\end{equation}

For Ground ($l_w$) Nodes:
\begin{equation}
g_{n_{\text{ground}}} = g_{s}[s] + h(c, n) + c_{n}
\end{equation}

Where the equations work based on the cost map provided before and the following values are considered :
\begin{itemize}
    \item $g_{n_{\text{aerial}}}$ is the cost of node $n$ for aerial traversal.
    \item $g_{n_{\text{ground}}}$ is the cost of node $n$ for ground traversal.
    \item $g_{s}[s]$ is the current cost to reach the start node $s$.
    \item $h(c, n)$ is the heuristic cost from node $c$ to node $n$.
    \item $c_{n}$ is the cost associated with node $n$ in the costmap.
    \item $\Delta e$ is the rounded absolute difference in elevation between nodes $c$ and $n$, normalized by resolution $\text{res}$.
    \item $Cle_{a}$ is an additional cost factor for aerial traversal.
    \item $\text{tup}(n)$ is a tuple representing node $n$.
\end{itemize}

\subsubsection{Grid-based Discretization}
The uniform grid-based discretization is already created while doing mapping which is further used to create a graph by generating a square grid with equidistant vertices. Also for path planning another 2D grid layer with elevation values and a cost map are used in the path planning which reduces the computation expense for the path planning compared to 3d cube layers and allows for near real-time performance from the planner.

Moreover, the path planning is calculated at the level of the robot's center of mass. Thus, the robot's geometry and safety distance are considered and clearance is added to the non-traversable area or flying areas.

\section{Post-Processing Optimal Path}

After creating the set of waypoints, $X = \{x_0, x_2,...,x_N\}$, using the modified A$^\star$ algorithm, a post-process of the path is realized in order to simplify the path waypoints and check collision. It is done in two steps; first, the path is pruned by removing all the useless waypoints which are collinear. Each of these actions is realized only in a partition of the path using the same mode of locomotion. This partition is described by the equation \ref{eq::partition}. Therefore, the transition waypoints do not change during this phase; only the path between two nodes is modified. 

\begin{equation}
\begin{aligned}
    X = \bigcup  X_i &= 
    \begin{cases}
     X_1 = \{x_i|0\leq i \leq k_1, \forall x_i \in M_{j1}\}\\
     ...\\
     X_n = \{x_i|k_{n-1}\leq i \leq k_n, \forall x_i \in M_{jn}\}\\
    \end{cases} \\
\end{aligned}
\label{eq::partition}
\end{equation}
\noindent where $M_{ji}$ represents the locomotion mode associated with the partition $X_i$.
\subsection{Path Pruning}

The Path Pruning Algorithm streamlines trajectories by iteratively removing redundant waypoints from a provided path\cite{pathprunning}. Given a partition $X = [x_0, ... x_N]$ of the global path, the algorithm assesses colinearity between waypoints $x_i$, $x_{i+1}$, and $x$. If colinear, $x_i$ is deleted to simplify the path. When not colinear, the algorithm checks edge clearance; if obstructed, intermediate waypoints between $x_i$ and $x$ are removed. This process ensures an efficient, safer, and smoother trajectory, benefiting robotic navigation and resource utilization. The pruned partition $X$ is then returned, offering an optimized path for autonomous systems.

\begin{algorithm}[H]
\caption{Path Pruning Algorithm}\label{alg::path_pruning}
\KwIn{$X = [x_0, ... x_N]$ partition of the global path}
\KwOut{$X$ waypoints of the pruned partition}

$i \gets N - 1$\;
$x \gets x_N$\;
\While {$i \neq 0$}{ 

    \If {Waypoints $x_{i+1}$ and $x_i$ are colinear with $x$}{
        $delete\_node(x_i)$\;
    }
    \ElseIf {$clear\_edge(x_i, x) \neq True$}{
        $delete\_nodes\_between(x_{i+1}, x)$\;
        $x = x_i+1$\;
    }
    $i \gets i - 1$\;
}
\Return{$X$}
\end{algorithm}
\medskip

\medskip

\begin{figure}[H]
    \centering
    \includegraphics[width=0.8\textwidth]{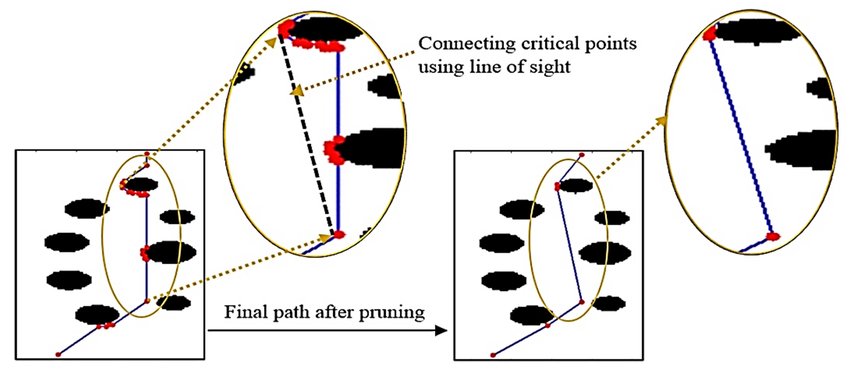}
    \caption{Example of path Pruning \cite{pathprunning}.} 
    \label{fig::path_prune}
\end{figure}

A simplified example of the result of this process is presented in Figure \ref{fig::path_prune}.

\subsection{Multimodal Locomotion mode selector and path follower}
\label{sec::Path planner}

The 3D path planner generates a comprehensive trajectory, complete with locomotion flags, which is then received by the subsequent node. This node efficiently dissects the trajectory, intelligently segmenting it based on changes in locomotion flags. For instance, upon encountering a transition to ground locomotion, the specific ground path is meticulously extracted and promptly relayed to the ground path follower.

\begin{algorithm}
\SetKwInOut{Input}{Input}
\SetKwInOut{Output}{Output}
\Input{3D waypoints with locomotion flags}
\Output{Comprehensive log indicating path completion}
\caption{Multimodal Robot Waypoint Traversal}

\While{not all waypoints completed}{
    Get the current waypoint's locomotion flag: $flag$\;
    
    \If{$flag$ is ground locomotion}{
        Publish ground path segment\;
        Activate ground path follower\;
        Execute P-control for differential drive\;
        Monitor progress towards ground segment's endpoint\;
        
        \If{ground segment endpoint reached}{
            Switch to next waypoint with different $flag$\;
        }
    }
    \ElseIf{$flag$ is aerial locomotion}{
        Publish aerial path segment\;
        Initiate aerial path follower using mavros and mavlink\;
        Monitor and control trajectory using ardupilot\;
        Track progress towards aerial segment's endpoint\;
        
        \If{aerial segment endpoint reached}{
            Switch to next waypoint with different $flag$\;
        }
    }
}

Generate a comprehensive log indicating path completion\;
\end{algorithm}

The ground path follower employs a straightforward yet effective p-control motion planner, masterfully orchestrating command velocities tailored for the robot's differential drive. Throughout this process, a vigilant monitoring system ensures the attainment of each segment's destination. The moment the first segment culminates in success, a seamless transition to the subsequent phase transpires.

Upon detecting the shift to aerial locomotion, the planner promptly assembles the aerial trajectory, which is then seamlessly handed over to the aerial path follower. The seamless integration of Mavros and mavlink facilitates the transfer to ardupilot, guiding the robot along the specified aerial trajectory with precision and finesse. As the robot diligently navigates the trajectory, a pivotal signal is relayed upon reaching the conclusion of each segment, acting as a catalyst for the commencement of the ensuing trajectory compilation.

This intricate orchestration continues, ensuring the robot gracefully navigates through each trajectory segment, executing the prescribed locomotion mode. At every turn, the system adeptly adjusts, dictating input commands to the hip joints of m4, ensuring a seamless transition between aerial and ground locomotion modes.

This elaborate process perseveres until no trajectory segments remain, concluding with a comprehensive log signaling the triumphant completion of the path. It is worth noting that the system consistently monitors and responds to the dynamic locomotion demands, all while ensuring the robot's movement aligns flawlessly with the designated path.

\section{Costs for energy calculation}

\begin{figure}[H]
    \centering
    \includegraphics[width=1\textwidth]{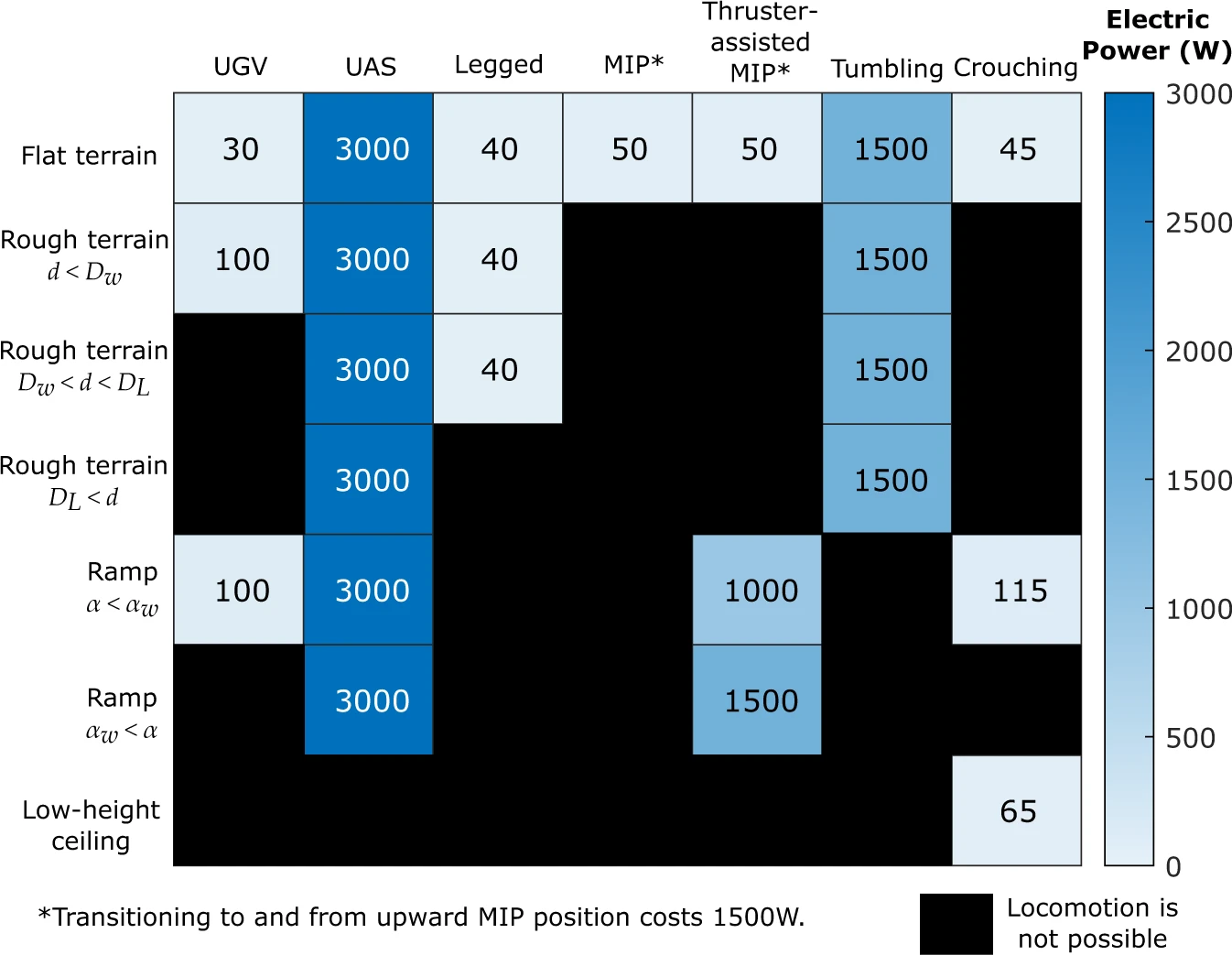}
    \caption{Locomotion modes versus electrical power consumption\cite{M4_nature}}
    \label{fig::energy}
\end{figure}

As explained in section \ref{subsec::Path planner}, path planning aims to optimize the total amount of energy used during the entire course to increase the robot's autonomy efficiency, and durability. To compute the real cost of the robot following formulas are used while assuming that the velocity of the robot is constant. Also, the figure \ref{fig::energy} tells us the real cost of Locomotion modes versus electrical power consumptions.

\subsection{Cost of Ground Wheeled Mobility}

The driving cost of 1 meter is computed with the average consumption power in driving $P_d$, and the robot's velocity in driving locomotion $v_d$.

\begin{equation}
    C_d = \frac{P_d}{v_d}
\end{equation}

\subsection{Aerial Mobility Cost}

In an analog way, the flying cost to move 1 meter is computed with the average flying power consumption $P_f$, and the flying velocity $v_f$.

\begin{equation}
    C_f = \frac{P_f}{v_f} 
\end{equation}

\subsection{Transformation Costs}

To change from one mode to another the drone has to do a transformation of its morphology. The transformation requires some energy but also some time, so it's necessary to associate a cost to it. Therefore, the energy consumed is used based on the servomotor electrical power consumption, $P_t$, during the transition and the transformation time $t_t$.

\begin{equation}
    C_t =\int_{0}^{t_t} P_t(\tau) d\tau 
\end{equation}

% --- EOF ---

% testing and discussion
\chapter{Testing and Results}
\label{chap:testing}

In this section, the outcomes of the implemented simulation in Gazebo are presented, followed by an analysis of the elevation mapping process. Subsequently, the testing scenarios for the developed path planner are described, accompanied by relevant material.

\section{Simulation in Gazebo} 

The simulation environment was established in Gazebo to replicate real-world conditions for path planning and navigation testing. The simulated environment included realistic terrain models, sensor models, and robot dynamics. The implemented simulation allowed for the evaluation of the path planner under various scenarios and conditions.
\begin{figure}[htbp]
    \centering
    \begin{subfigure}{0.45\textwidth}
        \centering
        \includegraphics[width=\linewidth]{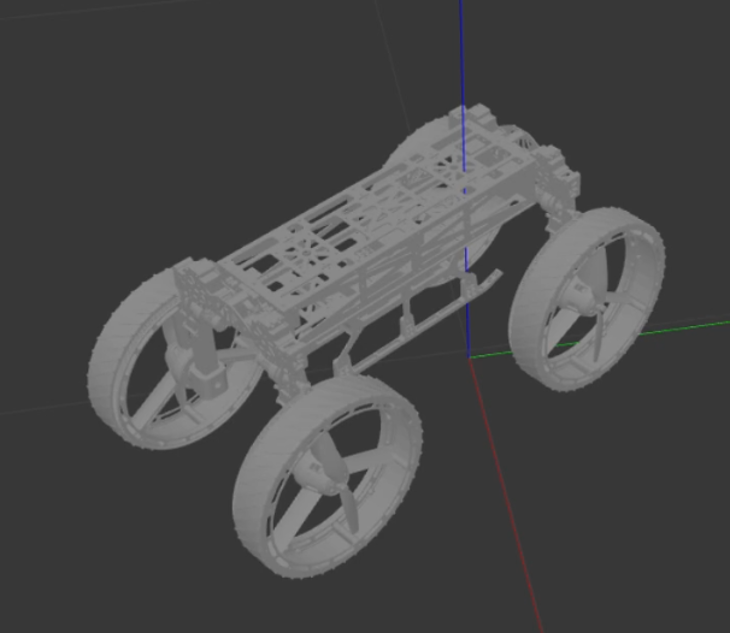}
        \caption{Ground mode M4}
        \label{fig:image1}
    \end{subfigure}
    \hfill
    \begin{subfigure}{0.45\textwidth}
        \centering
        \includegraphics[width=\linewidth]{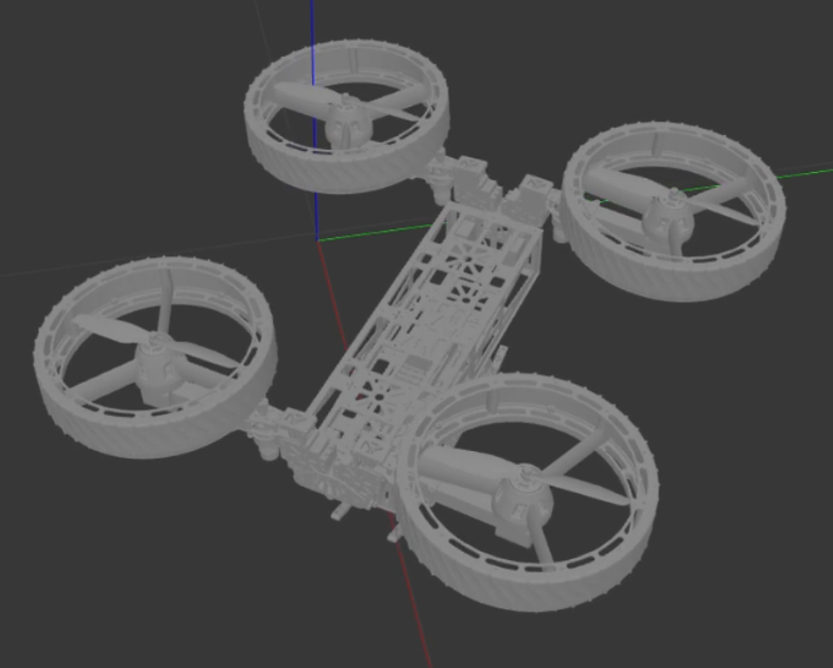}
        \caption{Aerial mode M4}
        \label{fig:image2}
    \end{subfigure}
    \caption{The above diagram shows the simulated model in the gazebo of M4}
    \label{fig:side_by_side}
\end{figure}
the above figure \ref{fig:side_by_side} shows the M4 robot in the URDF mode in the gazebo while also showing both morphing capabilities of the robot.
One of the things for the morphing was to tune the PID control of all 4 joints for simulation which took a lot of time.
also, we connected another Realsense depth camera at the bottom of the robot in simulation as the mapping area around the robot while M4 is in flight mode was much easier and gave an idea to implement a camera on the hardware M4 so that it can map while its flying using its downward-facing camera and the front-facing camera would work when the M4 is in wheel locomotion mode.

\section{Elevation Mapping}

\begin{figure}[H]
    \centering
    \includegraphics[width=1\textwidth]{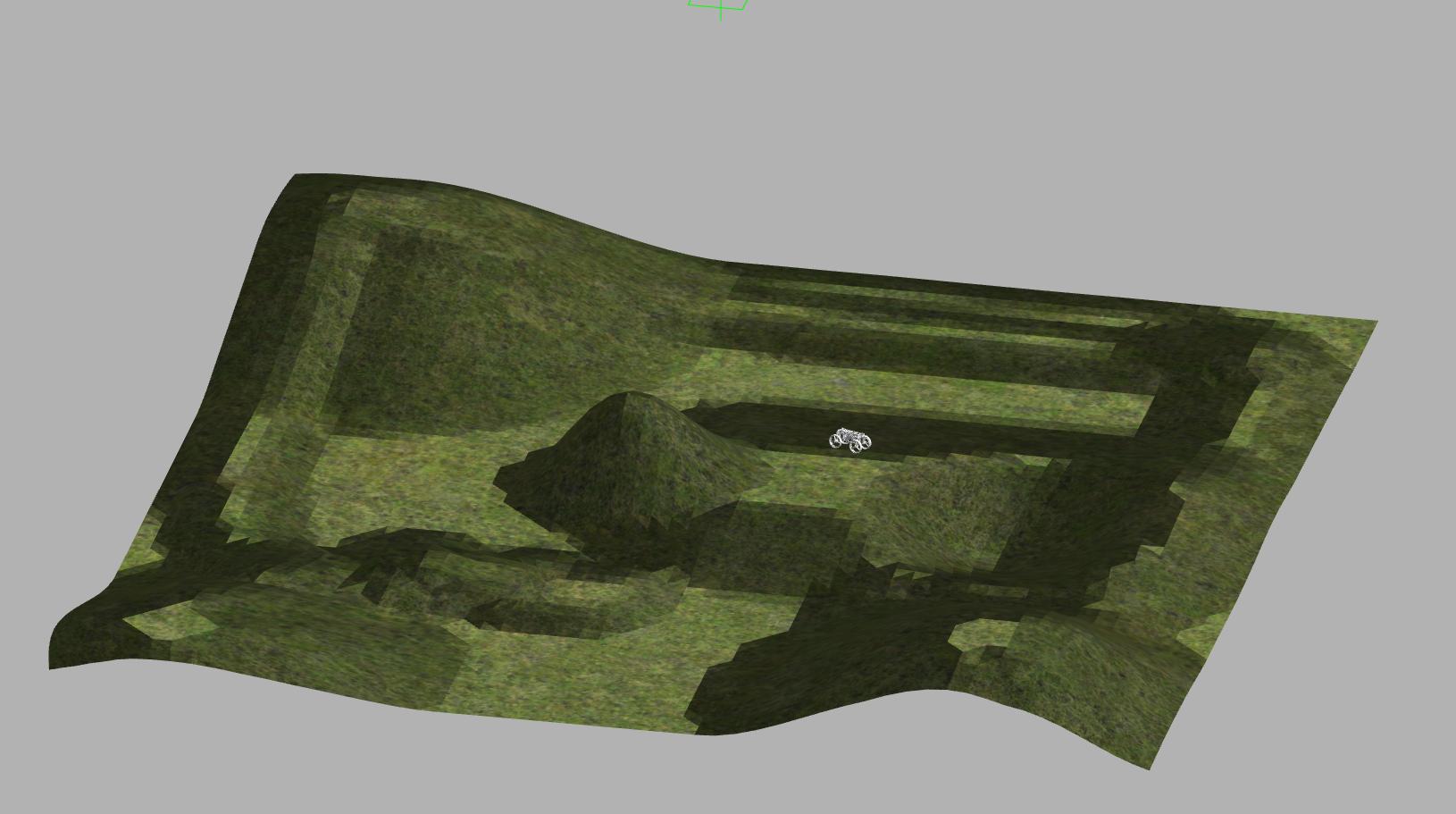}
    \caption{Uneven surfaces in gazebo simulation environment created using blender}
    \label{fig::gazebo_uneven}
\end{figure}

\begin{figure}[htbp]
    \centering
    \begin{subfigure}{0.45\textwidth}
        \centering
        \includegraphics[width=\linewidth]{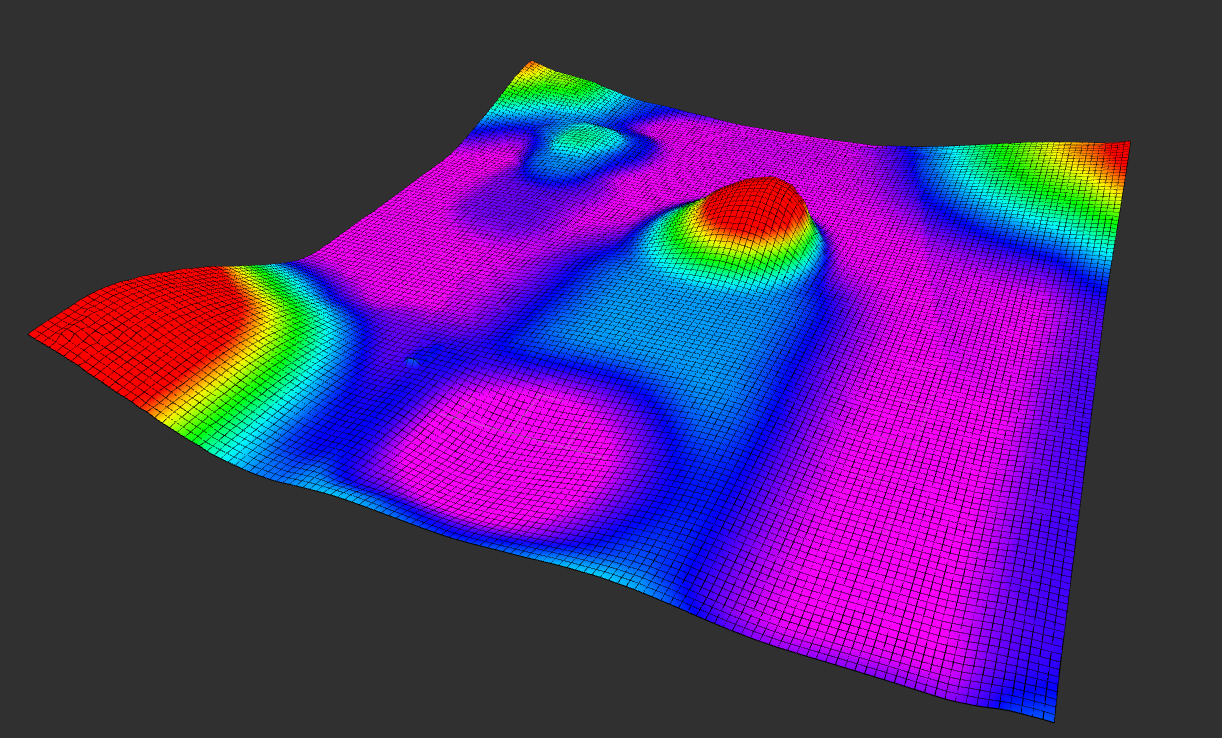}
        \caption{elevation map with traversability from zero to one while red color shows zero and purple shows one}
        \label{fig:elevation_c}
    \end{subfigure}
    \hfill
    \begin{subfigure}{0.45\textwidth}
        \centering
        \includegraphics[width=\linewidth]{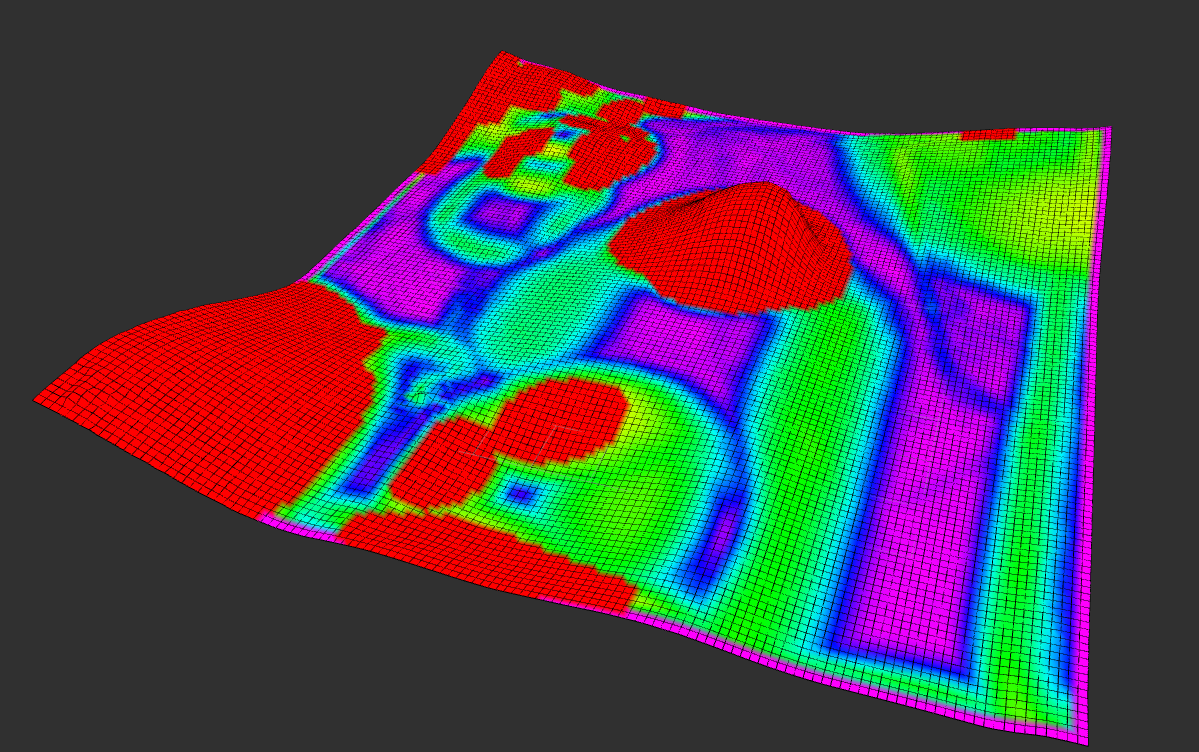}
        \caption{elevation map with multimodal cost map from zero to sixty while red color shows sixty and purple shows one}
        \label{fig:costmap}
    \end{subfigure}
    \caption{Output of elevation mapping and costmap creation}
    \label{fig:elevation_gazebo}
\end{figure}

Elevation mapping is a critical aspect of path planning in outdoor environments, as it influences the robot's ability to traverse uneven terrain effectively. The elevation mapping algorithm was applied to generate an accurate representation of the terrain's vertical profile. By fusing data from range sensors, the algorithm produced a comprehensive elevation map and traversability map\ref{fig:elevation_gazebo}from the environment in figure \ref{fig::gazebo_uneven}.

\section{Testing Scenarios}
\label{chap:testing1}
To evaluate the performance of the developed path planner, three distinct testing scenarios were designed, each representing a different navigational challenge. The scenarios aimed to assess the planner's adaptability and efficiency across varying terrains and obstacles.
The first hilly terrain is chosen to test to check whether our path planner follows the most optimal path in uneven terrain in single locomotion when it knows that going over slop or bump costs more than going around it.
For the other quantitative analysis, we chose 3 edge cases where a Multimodal path planner will be required and a simple planner will not work to operate multimodal robots.

\subsection{Hilly Terrain} 

\begin{figure}[htbp]
    \centering
    \begin{subfigure}{0.45\textwidth}
        \centering
        \includegraphics[width=\linewidth]{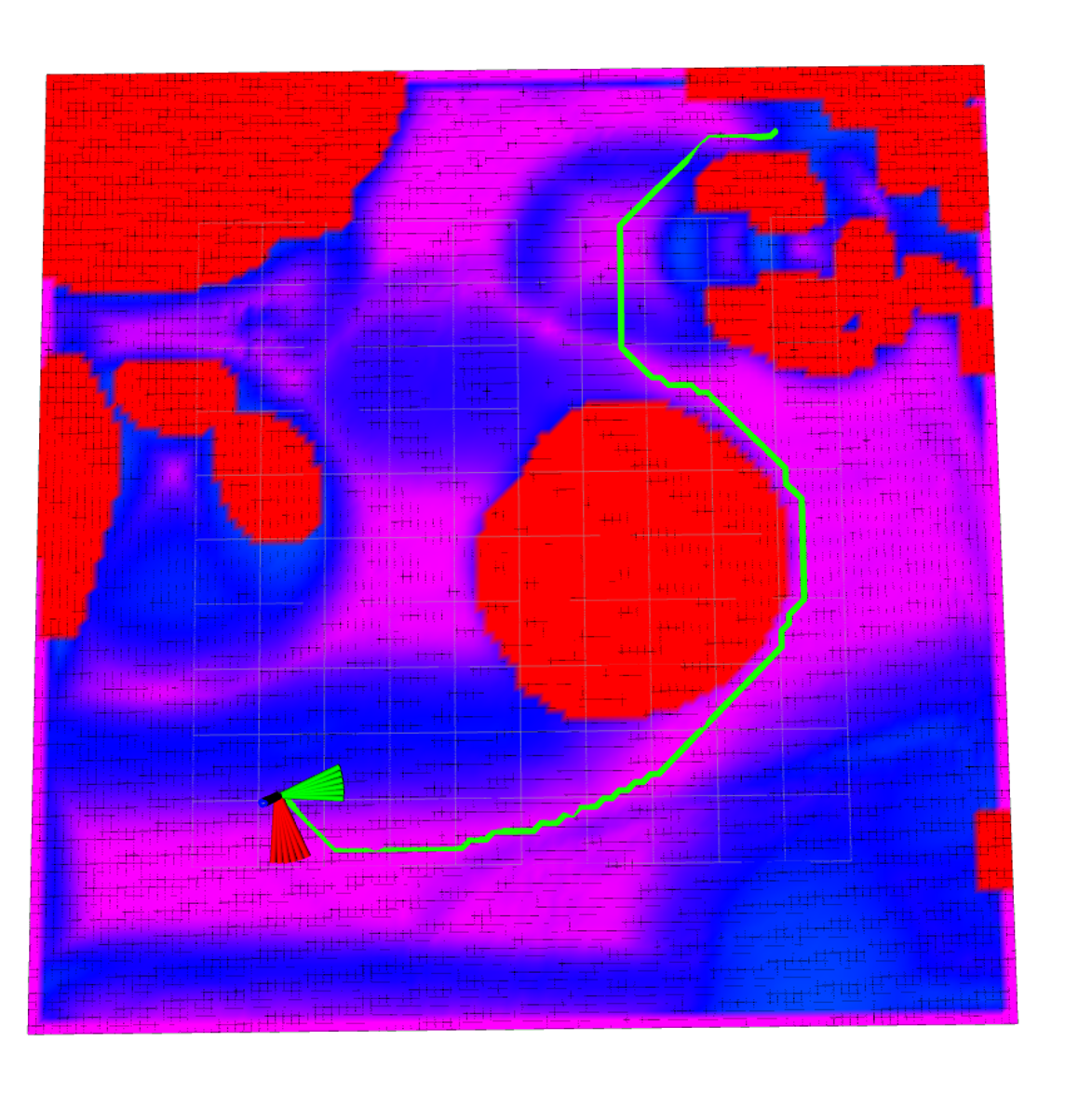}
    \end{subfigure}
    \hfill
    \begin{subfigure}{0.45\textwidth}
        \centering
        \includegraphics[width=\linewidth]{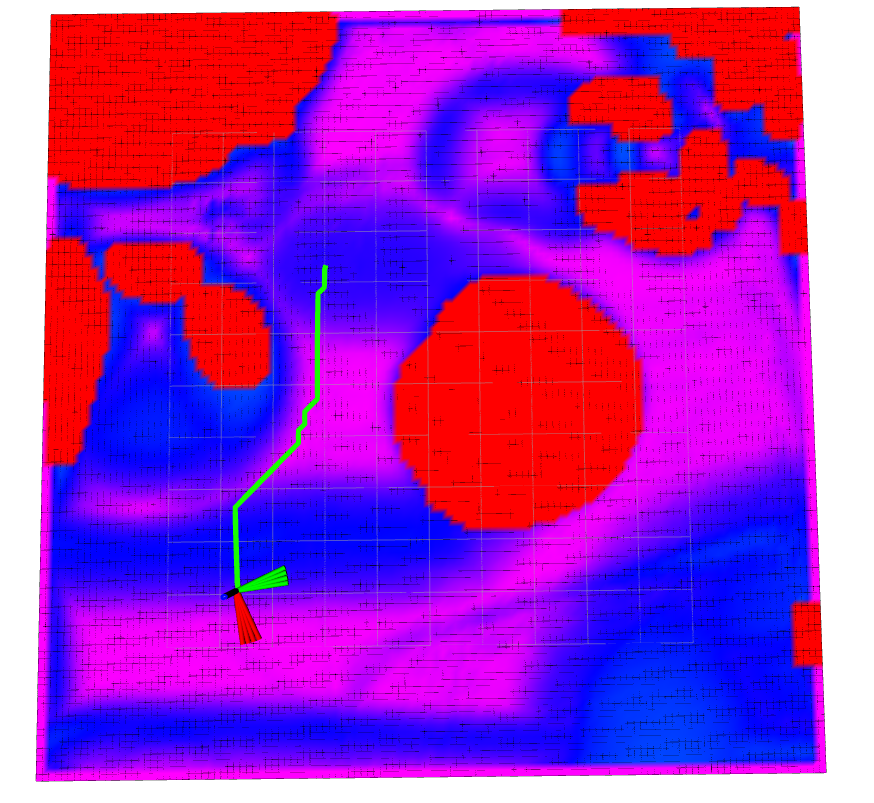}
    \end{subfigure}
    \caption{Path planned in uneven terrain with different end goals}
    \label{fig::pathu1}
\end{figure}

This scenario involved navigating through hilly terrain \ref{fig::gazebo_uneven} with varying degrees of slope. This test aimed to evaluate the planner's capability to handle elevation changes and plan trajectories that minimized energy consumption. Figure \ref{fig::pathu1} illustrates the robot planning multiple paths for different goals in the costmap we have created. the image also shows the costmap where you can see purple is the most easily traversable area and red is the non-traversable area and accordingly the cost is placed from 0 to 60 for the costmap. from figure \ref{fig::pathu1} you can see that when selecting the very edge of the path it did not go through the slight slope that's covered in blue color but traversed it where it found a flat area. for the other path, we chose to give the planner a goal on the other side of the slop to test and the planner did optimize the path and directly headed for the goal as going around would have cost it more.

\subsection{Edge Test case 1 setup and evaluation}
\begin{figure}[H]
    \centering
    \includegraphics[width=0.7\textwidth]{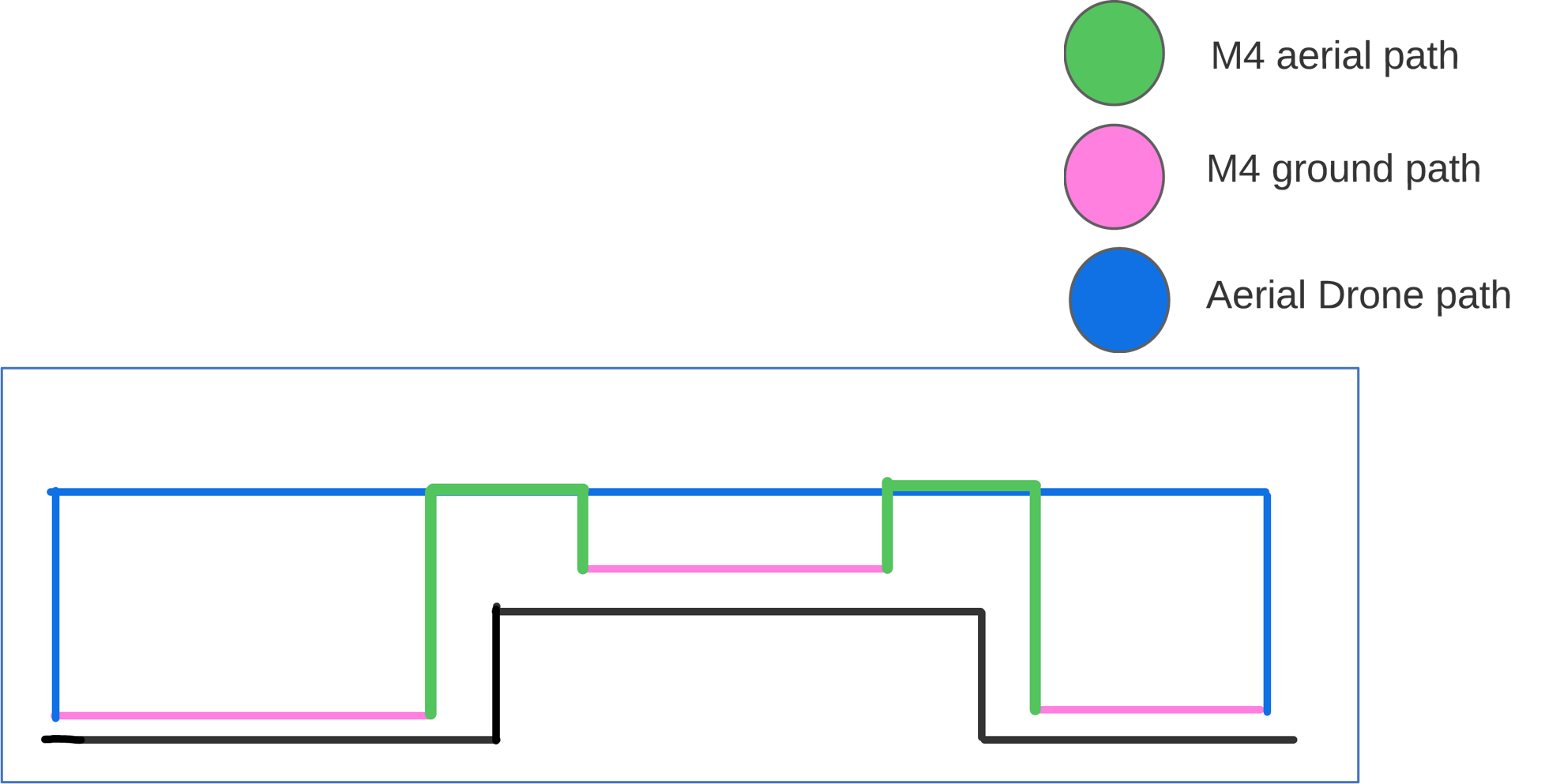}
    \caption{Sketched diagram of the path planned by M4 and Drone for test 1}
    \label{fig::1ta}
\end{figure}

This scenario tests the 3d path planner's ability to plan for a step environment where the ground heights are different and the robot has to plot a path to the goal with efficient energy using its multimodality. The environment may seem to have a way around it but it's actually designed to give the robot only a step environment and no flat surroundings as you can see in the mapped costmap figure \ref{fig::1tc}.
\begin{figure}[H]
    \centering
    \includegraphics[width=0.7\textwidth]{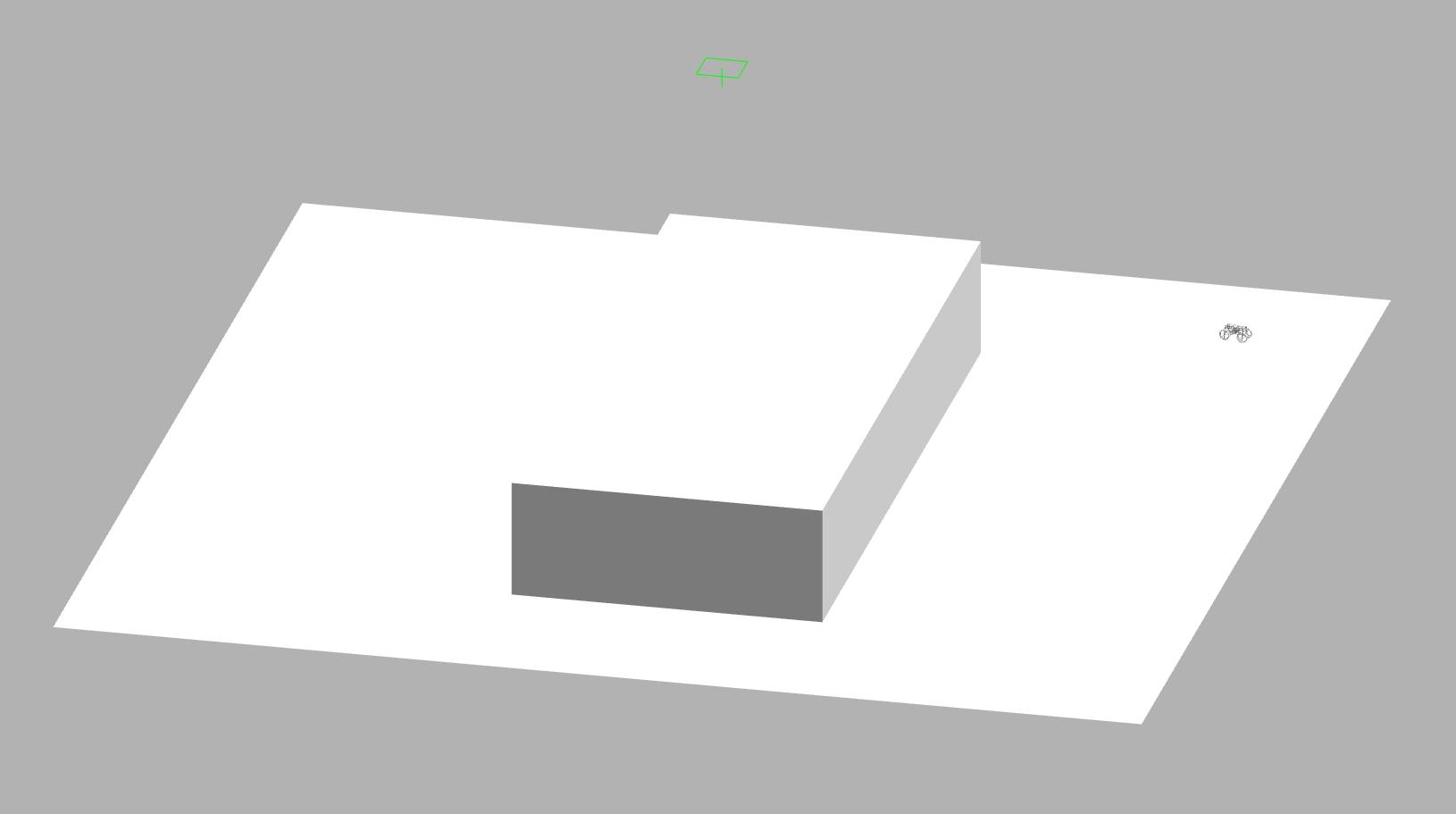}
    \caption{Environmental model of Scenario for Test case 1}
    \label{fig::1t}
\end{figure}
Figure \ref{fig::1ta} is a simplified sketch to show the path planned by our 3d planner and a path chosen by an aerial drone. In all coming cases, we have utilized a comparison module with an aerial robot which always starts from the start point and flies to avoid all obstacles, and then lands at the goal.

The color scheme for all sketched diagrams, where the color green in the path in figure \ref{fig::1ta} is to show the M4 robot's flight path and the purple color is to show the M4 robot's ground path in the path planned while the blue color indicates the path for the aerial robot.
The 3D path planner for M4 calculates the path in the step where it uses its wheel mobility to go until its wheel locomotion is not possible then it morphs into aerial mode and then takes off goes around the edge of the step and lands from where it can use its wheel mode as its more energy efficient, the whole detection of modes is done through traversability which indicates the robot which areas it can use for its wheel mobility. After landing on top of the step it goes until the end of the step and here you can see it sees a non-traversable area and calculates the elevation difference, morphs and takes off to have clearance with the edge, and then lands at the bottom of the step and converts into wheel locomotion mode and follows the rest of the path and then comes to a halt when the goal is reached.
The implementation of the path in 3d is seen in figure \ref{fig::1tc} where the costmap colorbar shows the traversable and non-traversable areas. 
\begin{figure}[H]
    \centering
    \includegraphics[width=0.7\textwidth]{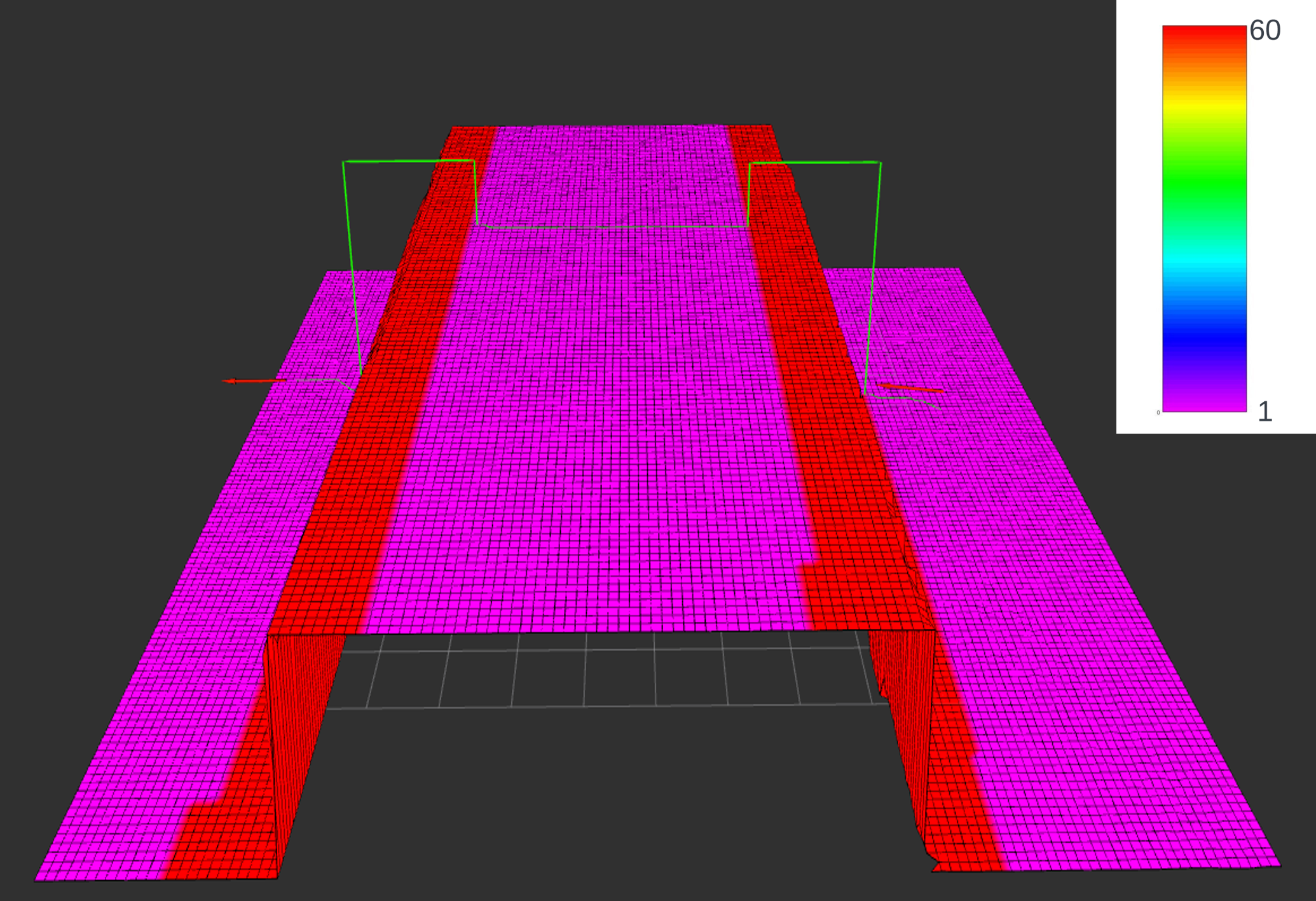}
    \caption{Multimodal Costmap and path planned based on environment \ref{fig::1t}}
    \label{fig::1tc}
\end{figure}

The detailed results are recorded in Table \ref{tab:performance}.

\subsection{Edge Test case 2 setup and evaluation}
\begin{figure}[H]
    \centering
    \includegraphics[width=0.65\textwidth]{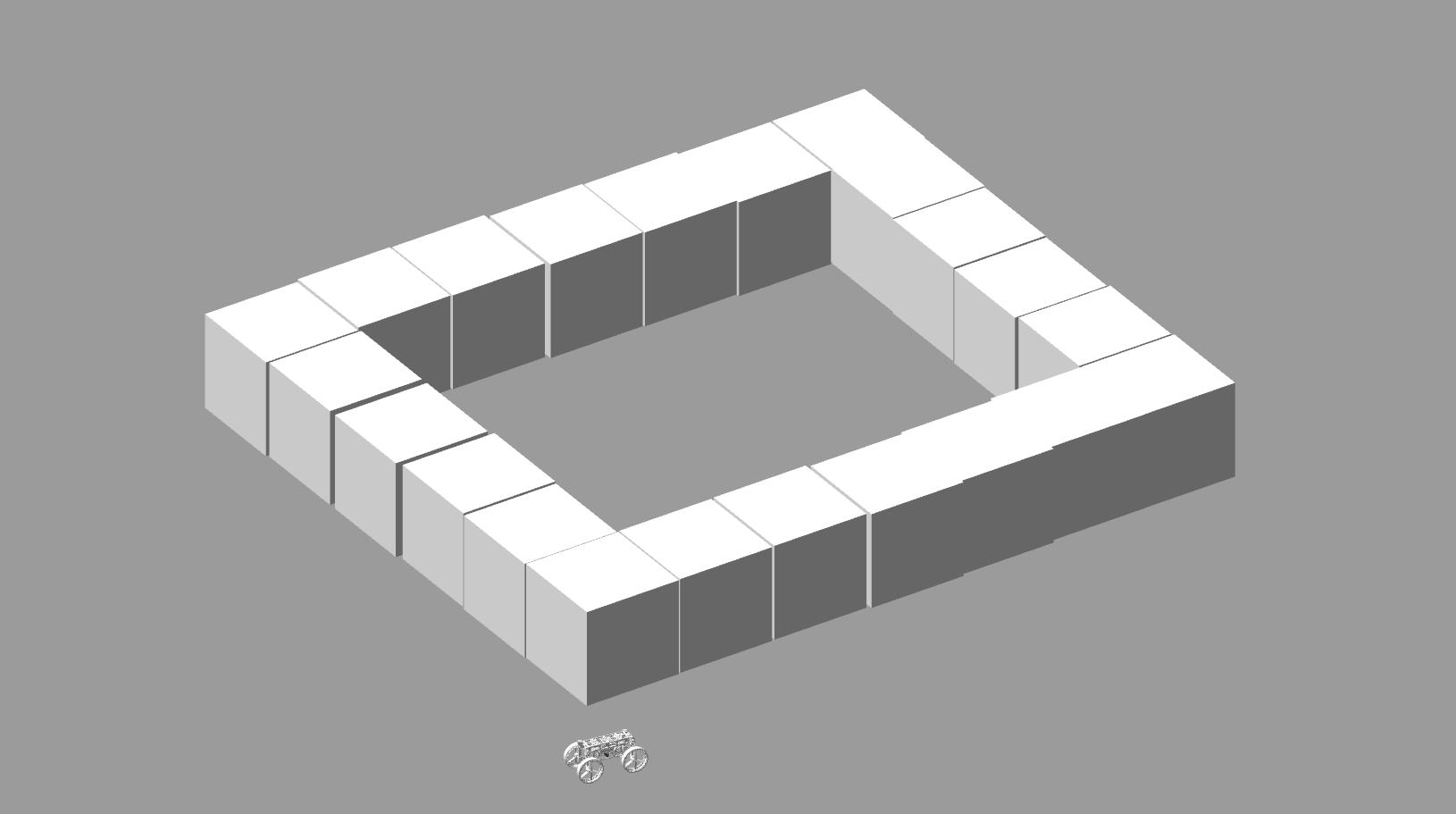}
    \caption{Environmental model of Scenario for Test case 2 }
    \label{fig::2t}
\end{figure}

\begin{figure}[H]
    \centering
    \includegraphics[width=0.8\textwidth]{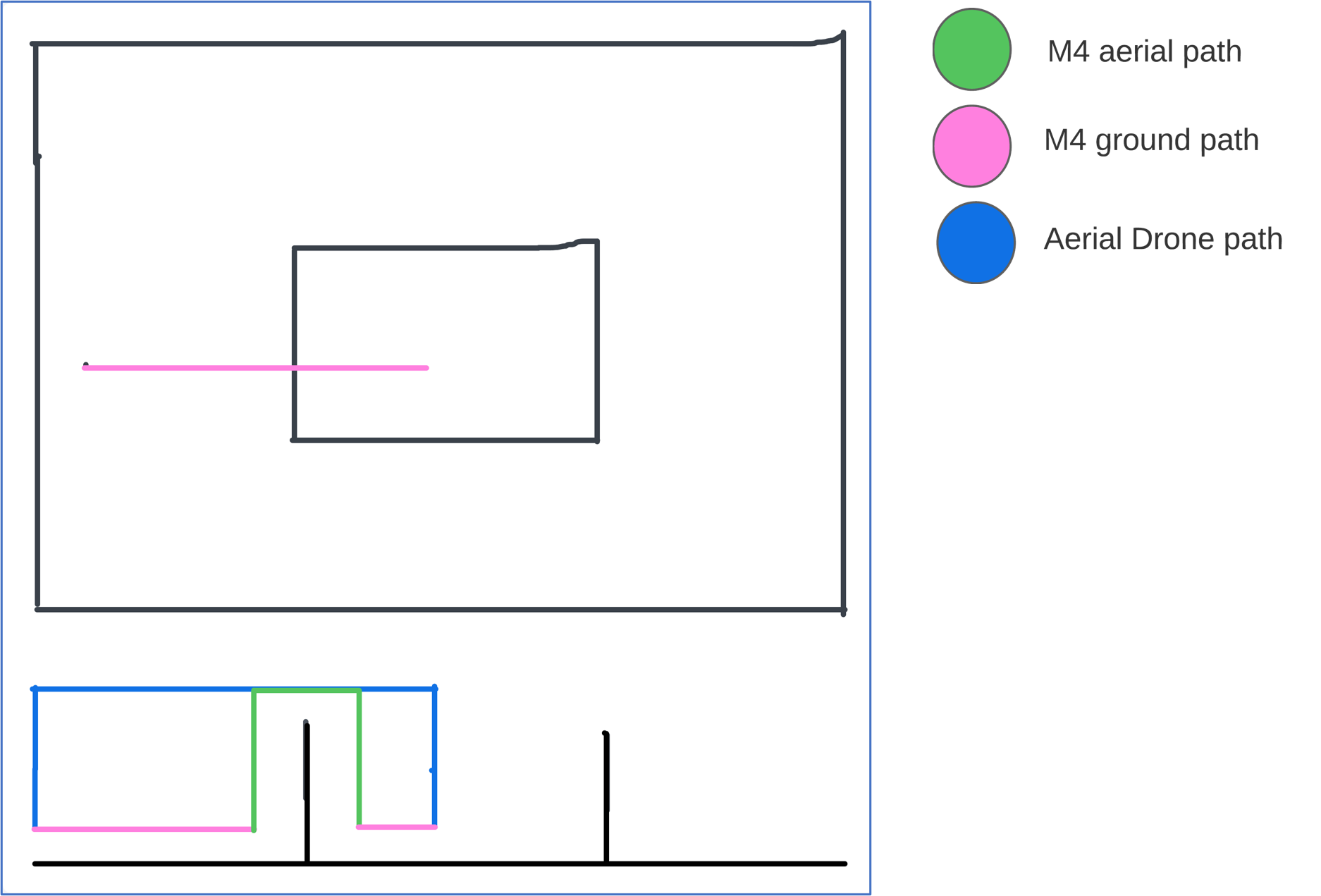}
    \caption{Sketched diagram of the path planned by M4 and Drone for test 2}
    \label{fig::2ta}
\end{figure}

\begin{figure}[H]
    \centering
    \includegraphics[width=0.7\textwidth]{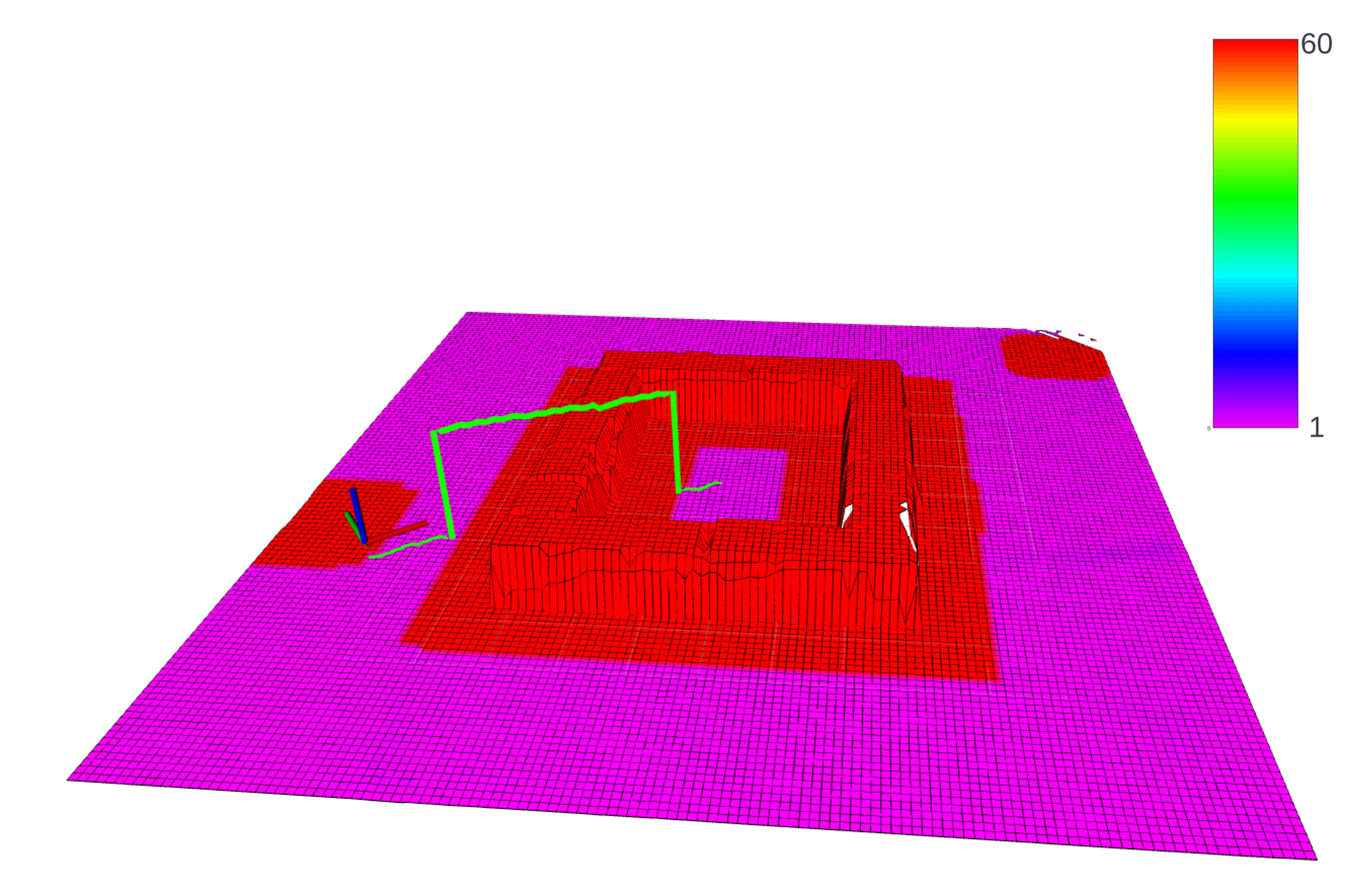}
    \caption{Multimodal Costmap and path planned based on environment \ref{fig::2t}}
    \label{fig::2tc}
\end{figure}

This scenario is chosen to show that where a single-wheel locomotion robot cannot reach, the Path is blocked to the goal from all sides and the M4 robot has to go to the endpoint.
Figure \ref{fig::2t} shows the environment used for the test case where we have put a square box with some gap in them but not enough for the robot to go through but in a square formation to create a non-reachable area using wheel locomotion.
The 3d multimodal path planner starts the path from its current position and the goal is given in the center of the square. Figure \ref{fig::2ta} shows a simplified version of the planned path to understand the path created by the Pipeline for the M4 system. Here the same color conventions are used as before, here using traversability we determine a costmap that determines the inaccessible area for the m4 using its every locomotion, then using a star it plans a path to the goal while considering both modes to optimize the energy.
The total path is shown in 3d in rviz shown in figure \ref{fig::2tc} where we can see the path plotted from the start position or current position of the robot from the goal position inside the square and also the costmap color coding is described in the colorbar on the side.
The detailed results are recorded in Table \ref{tab:performance}.

\subsection{Edge Test case 3 setup and evaluation}
\begin{figure}[H]
    \centering
    \includegraphics[width=0.7\textwidth]{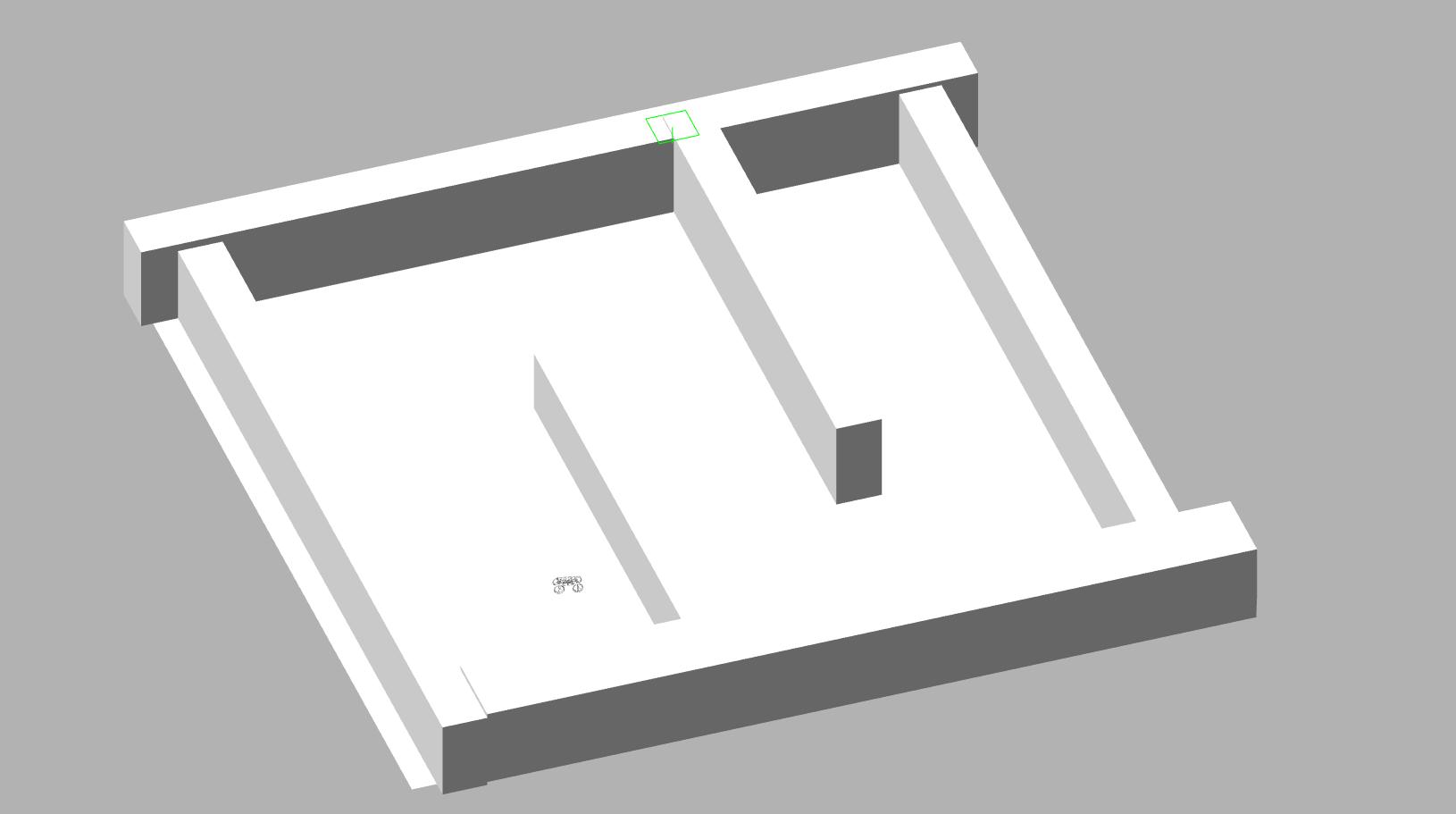}
    \caption{Environmental model of Scenario for Test case 3}
    \label{fig::3t}
\end{figure}

In this scenario, we aim to demonstrate the planner's reluctance to switch between locomotion modes unless absolutely necessary. The underlying principle is to emphasize the significance of energy conservation within the system, which might be overlooked when considering straightforward modes like multimodality or aerial drone navigation, as depicted in the illustrative Figure \ref{fig::3t}.

The simulated environment, as depicted in Figure \ref{fig::3t}, represents a maze-like configuration for the third test case. The primary objective here is to underscore the energy efficiency aspect of the M4 robot. While the destination is indeed reachable using wheel-based robot planners, our 3D path planner exhibits the ability to discern the most optimal path, even if it aligns with the trajectory planned for a mobile robot. This path is chosen based on the minimized cost to the goal, taking into account the energy factor.

Importantly, the planner doesn't adopt a different locomotion mode indiscriminately. Instead, it evaluates the energy efficiency of the current mode. If the current mode proves highly energy-efficient and the cost to the goal is comparatively lower, the planner maintains continuity in that mode rather than arbitrarily transitioning to another mode.

\begin{figure}[H]
    \centering
    \includegraphics[width=0.7\textwidth]{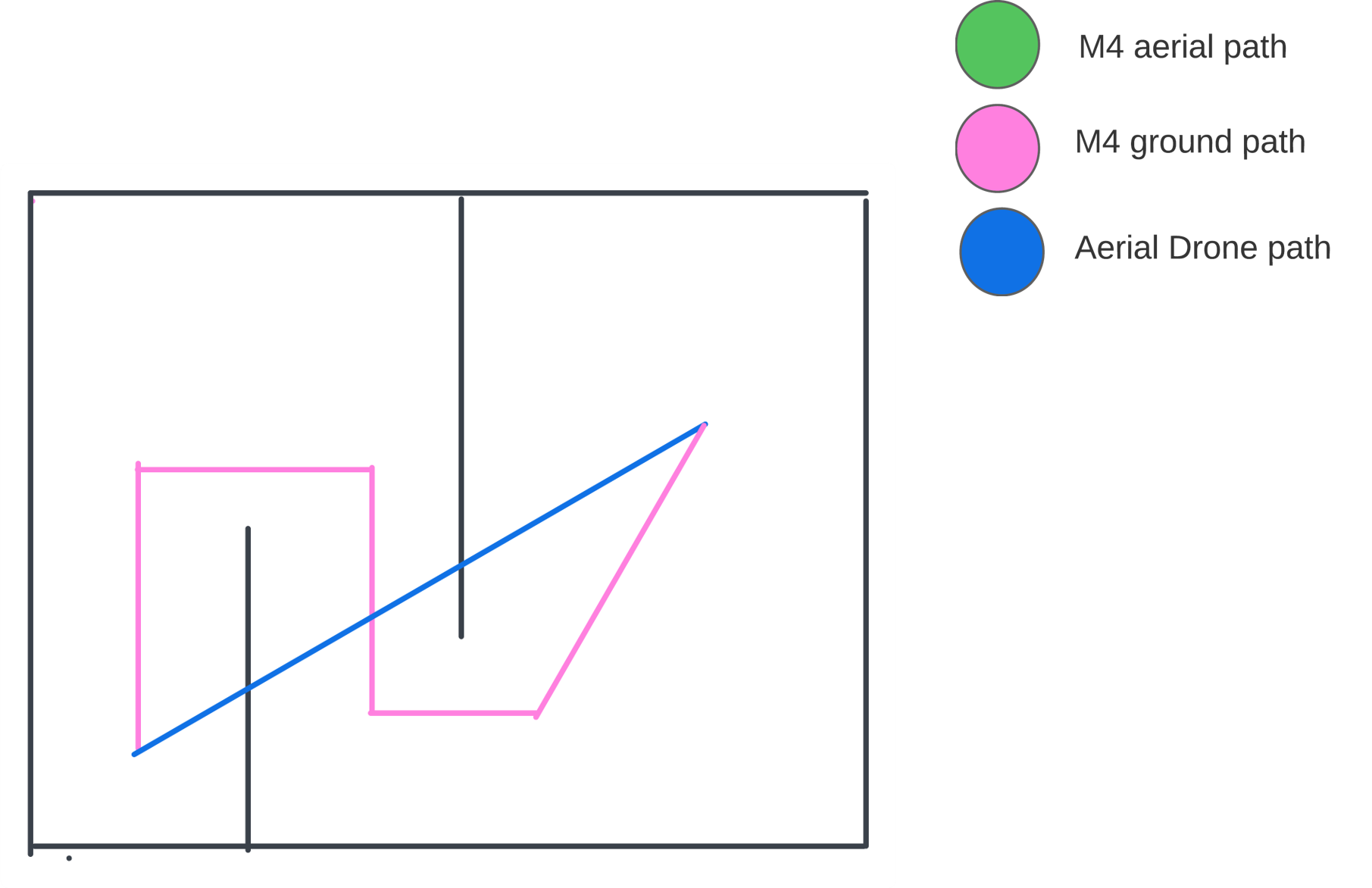}
    \caption{Sketched diagram of the path planned by M4 and Drone for test 3}
    \label{fig::3ta}
\end{figure}
\begin{figure}[H]
    \centering
    \includegraphics[width=0.7\textwidth]{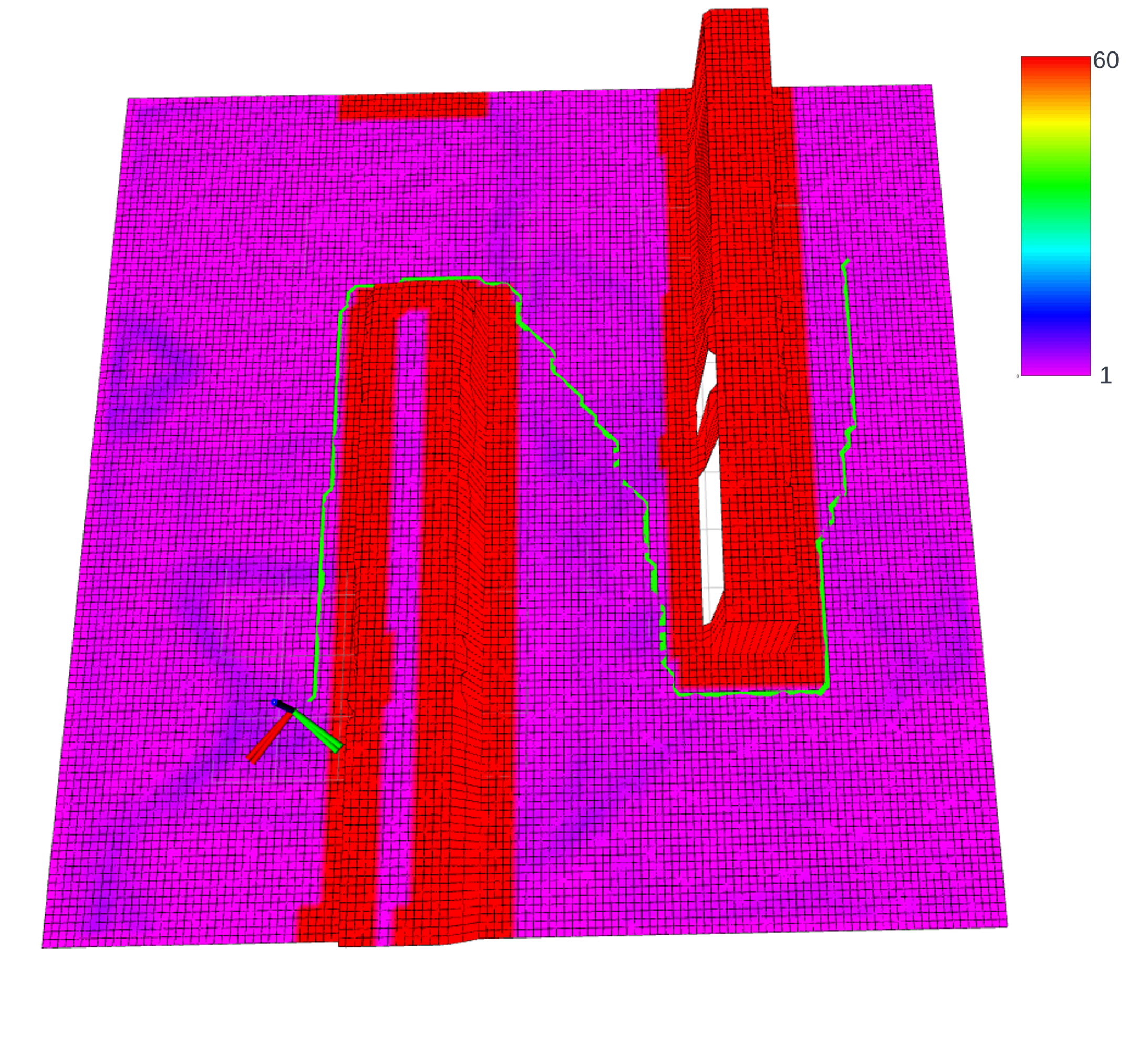}
    \caption{Multimodal Costmap and path planned based on environment\ref{fig::3t}}
    \label{fig::3tc}
\end{figure}


the sketch from figure \ref{fig::3ta} shows the path followed by M4 in ground mode while the Aerial drone directly goes to the end goal while checking the flight, here we assumed that the max height of the obstacle would be the altitude for the aerial drone. the 3D View of the planned path and the costmap can be seen in \ref{fig::3tc} which is taken from Rviz.
 
 The interesting fact about this environment is that in the determination of the 3d path, there is also a small traversable area on top of the 1st obstacle where our robot could land but it ignores it and follows the optimized path.

\subsection{Quantitative Analysis} 

Quantitative metrics were employed to assess the path planner's performance across the three testing scenarios. These metrics included path length, traversal time, and energy consumption. The planner's efficiency and effectiveness were compared across scenarios to ascertain its robustness and adaptability.
For the purpose of effective comparison, the energy consumption of the drone and M4 robot for aerial travel is assumed to be the same. Additionally, the following assumptions are made:

\begin{itemize}
    \item Time required for morphing between aerial and ground modes is assumed to be 5 seconds.
    \item Travel speed for both aerial and ground modes is assumed to be 1 m/s.
    \item M4 energy cost for 1 m of travel:
    \begin{itemize}
        \item Aerial travel cost: 60 W
        \item Ground travel cost: 1 W
        \item Morphing cost: 30 W
    \end{itemize}
\end{itemize}

\begin{table}[htbp]
    \centering
    \caption{Performance Metrics for Test Case Scenarios}
    \label{tab:performance}
    \begin{tabular}{|c|c|c|c|c|c|c|}
        \hline
        \multirow{2}{*}{Metric} & \multicolumn{2}{c|}{Energy (Joules)} & \multicolumn{2}{c|}{Time (s)} & \multicolumn{2}{c|}{Path Length (m)} \\ \cline{2-7}
        & M4 & Drone & M4 & Drone & M4 & Drone \\ \hline
        Test Case 1 & 994 & 1151 & 21.67 & 19.18 & 42 & 20 \\ \hline
        Test Case 2 & 475 & 600 & 10.6 & 10 & 21 & 10 \\ \hline
        Test Case 3 & 60 & 734 & 29.37 & 20.22 & 30 & 21 \\ \hline
    \end{tabular}
\end{table}

from the above table, we can see that in all the multimodal locomotion-oriented scenarios our path planner optimizes the most energy using its multimodal locomotion modes. the energy reduced in test case 3 is only 60 Jules to the drone's 734 joules which proves the energy efficiency of the multimodal robot on following the optimum path, as our planner systematically uses the modes only to advance the energy efficiency of the robot.
The significant energy savings demonstrated across these arduous test scenarios serve as a testament to the Multimodal Robot's efficacy in real-world outdoor environments. It is important to emphasize that these scenarios, while individually demanding, collectively encapsulate the challenges inherent in outdoor robotic navigation. The Multimodal Robot's consistent energy optimization, even in difficult circumstances, suggests that its path planning capabilities can be effectively generalized to real-world scenarios, translating to tangible energy savings and enhanced performance.

Overall, the results of the simulation and testing scenarios indicate the promising performance of the developed pipeline in various outdoor environments. The Multimodal locomotion system successfully navigated through uneven terrain and complex obstacle courses, showcasing its potential for real-world applications in outdoor robotic navigation.

% conclusion and Future work
 \chapter{Conclusion}
\label{chap:conclusion}

In conclusion, this thesis stands as a testament to innovation and rigorous exploration, yielding tangible outcomes that significantly advance the autonomy of the M4 robot. The fusion of meticulous research, cutting-edge technologies, and well-crafted methodologies has yielded remarkable contributions that lay the groundwork for a future of versatile and self-sufficient multi-modal robotics.

The thesis has successfully delivered a comprehensive framework for enhancing the M4 robot's autonomy. The 3D path planner, a core pillar of this framework, has been meticulously designed and validated to optimize energy-efficient navigation. By translating complex environments into a 2.5D map representation, the planner empowers the M4 robot to make informed decisions in real time, effectively determining optimal paths through intricate terrains. Through simulations, the planner's efficacy has been underscored, setting the stage for practical implementation on the physical robot.

The thesis also introduces a bespoke perception and navigation pipeline tailored to the M4 robot's unique morphing and multi-modal capabilities. This pipeline seamlessly integrates elevation mapping, traversability assessment, and navigation, enabling the robot to adapt to dynamic environments and make autonomous decisions during mode transitions. The culmination of these efforts is an autonomous decision-making system that leverages deep learning models, empowering the M4 robot to intelligently morph between its walking and flying modes based on real-time terrain analysis. This achievement not only exemplifies the potential of autonomous multi-modal robotics but also provides a stepping stone for further refinement and expansion to accommodate a broader array of locomotion modes.

The creation of a robust Gazebo simulation environment represents a significant breakthrough, offering a controlled platform for testing and validating the proposed autonomy algorithms. Through these simulated trials, the thesis demonstrates the M4 robot's ability to traverse challenging landscapes, seamlessly switching between modes and optimizing paths. This simulation-driven approach has been crucial in refining the autonomy algorithms, ensuring their readiness for real-world deployment.

In essence, this thesis demonstrates how the M4 robot can autonomously navigate complex terrains, leveraging its morphing and multi-modal capabilities to adapt and make decisions. By combining advanced path planning techniques, custom perception and navigation pipelines, and state-of-the-art deep learning models, the thesis has achieved outcomes that pave the way for a new era of autonomous robotics. As the M4 robot takes its digital steps in the simulated landscapes, it brings us closer to a future where robotic explorers autonomously traverse diverse terrains, ushering in a new age of intelligent and adaptable machines.
\section{Future Work}

The achievements and insights gained from this thesis lay a solid foundation for an exciting array of future explorations and advancements in the realm of autonomous multi-modal robotics. Building upon the outcomes and methodologies presented here, the following avenues offer compelling opportunities for further research and development:

\begin{enumerate}
    \item \textbf{Real-Robot Deployment and Field Testing:} The natural progression from simulation to real-world implementation presents an exciting opportunity to validate the proposed autonomy framework on the physical M4 robot. Real-robot testing will provide valuable insights into the system's performance under actual conditions and enable the assessment of its adaptability to dynamic and unpredictable terrains.
    
    \item \textbf{Enhanced Locomotion Mode Selection:} While the current work focuses on morphing between walking and flying modes, extending this to encompass all modes of the M4 robot opens doors to a more versatile and capable autonomous system. Training a custom deep learning locomotion mode selector for all modes of the M4 in a Gazebo environment would provide a comprehensive solution for mode transitions, further enhancing the robot's adaptability.
    
    \item \textbf{Semantic Segmentation for Enhanced Perception:} The integration of semantic segmentation into the mapping layer holds immense potential for expanding the robot's perception capabilities. By incorporating the ability to detect water bodies and surface properties, the M4 robot can make more informed decisions during navigation, ensuring safer and more efficient traversal through diverse environments.
    
    \item \textbf{Advanced Path Optimization:} The addition of a local planner to the existing autonomy framework presents a promising avenue for fine-tuning path optimization. By addressing dynamic obstacles and real-time constraints, the system can further enhance its ability to navigate complex terrains while considering energy consumption and mode transitions.
    
    \item \textbf{Multi-Robot Collaboration:} Extending the autonomy framework to support multi-robot collaboration opens up avenues for coordinated exploration and resource sharing. Investigating methods for M4 robots to communicate, share information, and collaborate intelligently can significantly amplify their efficiency and effectiveness in diverse scenarios.
\end{enumerate}

In essence, the future scope outlined by these points embodies a holistic approach toward creating a truly autonomous and versatile Multi-modal robotic system. By addressing challenges related to real-world deployment, perception enhancement, advanced planning, and collaborative capabilities, the M4 robot can evolve into a powerful tool for exploration, research, and practical applications across a spectrum of domains. The collective pursuit of these avenues promises to unlock the full potential of autonomous multi-modal robotics, bridging the gap between innovation and tangible impact.

% --- Bibliography ----
\bibliographystyle{IEEEtran}  %'plain' for standard, 'unsrt' for correct order

% include bibliography definition
\bibliography{bib/thesis}

% --- Appendix ---
\appendix
%include anything you need in the appendix
%\include{tex/appendixA}

% --- Index ----
%\printindex

% --- that's it ---
\end{document}